\documentclass[10pt,twocolumn,letterpaper]{article}

\usepackage[pagenumbers]{cvpr} 

\DeclareGraphicsExtensions{.png,.jpg,.pdf,.ai,.psd}
\definecolor{cvprblue}{rgb}{0.21,0.49,0.74}
\usepackage[pagebackref,breaklinks,colorlinks,allcolors=cvprblue]{hyperref}
\usepackage{gensymb}
\usepackage[export]{adjustbox}
\usepackage{overpic}
\usepackage{array}
\usepackage{graphicx}
\usepackage{xcolor}
\usepackage{ragged2e}
\usepackage{rotating}
\usepackage{multirow}
\usepackage[accsupp]{axessibility}
\def\paperID{42200} 
\def\confName{CVPR}
\def\confYear{2026}

\title{UniLight: A Unified Representation for Lighting}

\author{Zitian Zhang$^{1,2}$ \quad
Iliyan Georgiev$^2$ \quad
Michael Fischer$^2$ \\
Yannick Hold-Geoffroy$^2$ \quad
Jean-Fran\c{c}ois Lalonde$^1$ \quad
Valentin Deschaintre$^2$ \\
$^1$Université Laval, $^2$Adobe Research \\
}

\usepackage{algorithm}
\usepackage{algpseudocode}
\usepackage{array}

\newcommand{\method}{\textsc{UniLight}\xspace}
\newcommand{\myparagraph}[1]{\noindent\textbf{#1}}

\usepackage[nolist]{acronym}
\begin{acronym}
\acro{BRDF}{bidirectional reflectance distribution function}
\acro{HDR}{high-dynamic-range}
\acro{LDR}{low-dynamic-range}
\acro{NeRF}{Neural Radiance Field}
\acro{MAE}{masked auto-encoder}
\acro{PBR}{physically based rendering}
\acro{SAM}{Segment Anything Model}
\acro{ViT}{vision transformer}
\acro{VLM}{vision-language model}
\end{acronym}

\begin{document}


\makeatletter
\g@addto@macro\@maketitle{
    \vspace{-6mm}
    \centering
    \includegraphics{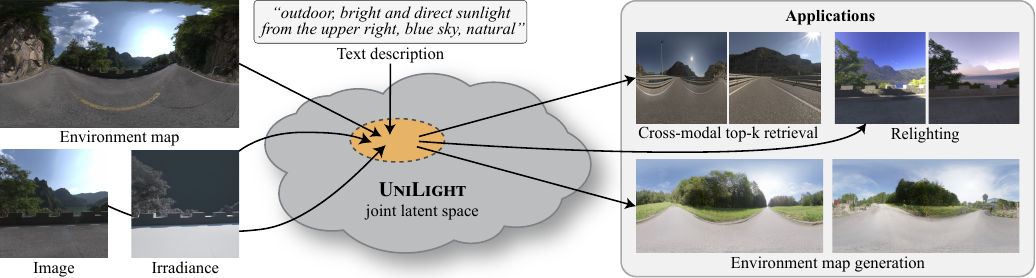}
    \vspace{-5mm}
    \captionof{figure}{
        We propose \method: A joint latent space unifying previously incompatible lighting representations. Our joint lighting embedding enables retrieval, example-based light control during image generation, and environment-map generation from various modalities.
    }
    \label{fig:Teaser}
    \vspace{4mm}
}
\makeatother

\maketitle


\begin{abstract}
Lighting has a strong influence on visual appearance, yet understanding and representing lighting in images remains notoriously difficult. Various lighting representations exist, such as environment maps, irradiance, spherical harmonics, or text, but they are incompatible, which limits cross-modal transfer. We thus propose \method, a joint latent space as lighting representation, that unifies multiple modalities within a shared embedding. Modality-specific encoders for text, images, irradiance, and environment maps are trained contrastively to align their representations, with an auxiliary spherical-harmonics prediction task reinforcing directional understanding. Our multi-modal data pipeline enables large-scale training and evaluation across three tasks: lighting-based retrieval, environment-map generation, and lighting control in diffusion-based image synthesis. Experiments show that our representation captures consistent and transferable lighting features, enabling flexible manipulation across modalities.\\
Project page: \small{\texttt{\url{https://lvsn.github.io/UniLight}}}.
\end{abstract}

\section{Introduction}
\label{sec:intro}

Lighting plays a central role in determining the visual appearance of images. Consequently, lighting representation and control are fundamental for image understanding, generation, and editing. However, lighting is inherently difficult to represent and can take many forms, including environment maps~\cite{DiffusionRenderer}, text descriptions~\cite{zhang2025scaling}, irradiance maps~\cite{zeng2024rgbx}, or simply reference images depicting example illumination~\cite{xing2025luminet}. Each representation has distinct, and often complementary, advantages and limitations, which makes them largely incompatible with one another. As a result, most lighting estimation or control methods are designed around a single representation and cannot easily adapt to others, limiting their flexibility~\cite{DiffusionRenderer,zeng2024rgbx}. 

Several works attempt to mitigate the limitations of the traditional lighting representations. For example, Li et al.\ \cite{li2020inverse} extend spherical Gaussians by estimating them locally over each pixel in the image, thus making them spatially-varying. More recently, inspired by \acp{NeRF} \cite{mildenhall2021nerf}, researchers have begun encoding lighting implicitly using a neural network \cite{boss2021neural,yao2022neilf,li2021neulf}. While these implicit representations can capture complex illumination effects, they are typically task-specific and are not compatible with other lighting representations.

In this work, we propose a joint latent space that unifies multiple lighting modalities (\cref{fig:Teaser}). We train modality-specific encoders for text, environment maps, irradiance maps, and images, to extract lighting information into a common high-dimensional embedding. Our encoders build on vision transformers (ViTs)~\cite{oquab2023dinov2} and text encoders~\cite{wang2025internvl3}, and are jointly trained with a contrastive learning objective across all modalities. To further improve light direction understanding, an auxiliary task predicts degree-3 spherical harmonics coefficients from the joint embedding.

We devise a multi-modal data pipeline to support training and to evaluate the learned representation on three downstream tasks: (a)~lighting-based retrieval, which retrieves modalities with similar illumination; (b)~environment-map generation, conditioned on any supported modality; and (c)~flexible lighting control for diffusion-based image synthesis~\cite{zeng2024rgbx}. We further ablate our design decisions and demonstrate that both the contrastive and auxiliary training objectives are critical to learn a latent space that captures accurate lighting structure and direction.

In summary, we make the following contributions:
\begin{itemize}[leftmargin=2em]
    \item A multi-modal data pipeline for unified lighting representation learning.
    \item A contrastive framework aligning the representation of lighting across text, image, irradiance, and environment map modalities.
    \item An auxiliary spherical harmonics prediction loss enhancing directional encoding.
    \item Extensive experiments demonstrating versatility across lighting-related applications.
\end{itemize}

\section{Related work}
\label{sec:RelatedWork}

\paragraph{Lighting modeling and estimation} have been long-standing challenges in computer vision and graphics. Early approaches often relied on simplified illumination models such as point or directional light sources to infer scene geometry from shading cues \cite{horn1970shape,woodham1980photometric}. In his seminal work, Debevec~\cite{debevec1998rendering} introduced the use of environment maps to capture real-world illumination, enabling realistic image-based lighting. Subsequent work extended this idea by estimating environment lighting from images using cues such as reflections, shadows, and geometry \cite{sato2003illumination,lalonde2014lighting,lalonde2012estimating}. 

Recent work has demonstrated strong performance in estimating lighting using learning-based methods, targeting indoor scenes \cite{Gardner_2019_ICCV,Gardner_2017_TOG,zhan2021emlight,song2019neural,li2023spatiotemporally}, outdoor environments \cite{hold2019deep,zhang2019all,zhu2021spatially}, or both \cite{legendre2019deeplight,dastjerdi2023everlight,Phongthawee2024DiffusionLight}. 
Contemporary approaches further leverage \acp{NeRF} \cite{yao2022neilf,rudnev2022nerf,jin2024neural_gaffer} or 3D Gaussian Splats \cite{bolduc2025GaSLight} to model high-frequency, spatially-varying illumination. 
For more details, we refer the reader to the comprehensive survey of \citet{einabadi2021deep} on light representations.

%
Recent approaches attempt implicit latent-space disentanglement, for instance using PCA \cite{haas2024discovering} or optimization \cite{li2024self,dalva2024noiseclr}. While these methods yield promising interpretable directions, they still require manual annotations. 
In this work, we propose to extract an implicit joint latent space that can serve as unified representation for disentangled lighting control in image generation.

\paragraph{Relighting} aims to modify the illumination conditions of an image while preserving the intrinsic identity of its elements. Traditional methods leverage inverse rendering techniques to estimate normals, materials, and lighting properties \cite{sato2003illumination,lombardi2015reflectance}, enabling re-illumination. These approaches, however, are typically sensitive to reconstruction accuracy: minor errors in geometry or material estimation can lead to implausible results, substantially reducing robustness. 

Portrait relighting has received particular attention \cite{zhou2019deep,sun2019single,nestmeyer2020learning,Hou_compose,mei2024holo,chaturvedi2025synthlight,ren2024relightful,zhang2025scaling,ponglertnapakorn2023difareli,mei2025lux} and attracted interest for digital avatars, virtual presence, and entertainment. 
In recent years, the relighting of generic scenes has gained traction \cite{kocsis2024lightit,griffiths2022outcast,magar2025lightlab,zeng2024dilightnet,scribble_light_choi_2025_cvpr,xing2025luminet}. 
In this context, the generation of shadows and coherent indirect illumination has emerged as a key challenge, with dedicated methods tackling shadow generation~\cite{sheng2022controllable,chadebec2025lbm} and removal~\cite{yoon2024generative,guo2023shadowformer}. 

In parallel with the rise of generative modeling, newer methods have begun to incorporate generative priors to achieve greater semantic consistency in relighting. For instance, latent-space editing in StyleGAN has been used for relighting~\cite{bhattad2024stylitgan}. More recent works apply diffusion-based models~\cite{SD3, rombach2022high} to relight by editing intrinsic image properties \cite{zeng2024rgbx,zhang2025zerocomp,lyu2025intrinsic} or by guiding lighting changes via user-specified cues such as shadows \cite{fortier2026spotlight}. For video and dynamic scenes, approaches leveraging diffusion models have even tackled video inverse-rendering, enabling relighting in the temporal domain \cite{DiffusionRenderer,he2025unirelight}. A related line of work estimates geometry first, then conditions a diffusion model on radiance hints to produce relit imagery \cite{erel2025practilight}. 

These methods are typically designed for a single type of lighting control, limiting users' ability to use their preferred lighting representation. We show that our joint lighting representation facilitates the design of multi-modal lighting applications for illumination control in both indoor and outdoor scenarios but also for environment-map generation or light-driven retrieval.
 

\begin{figure}
    \centering
    \includegraphics{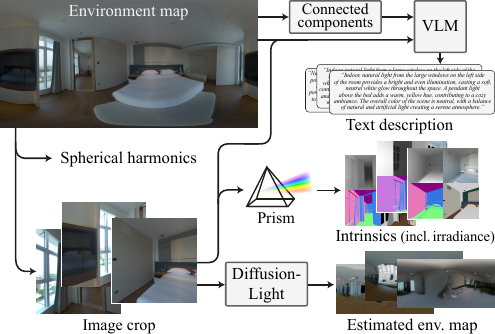}
    \vspace{-5mm}
    \caption{
        Dataset creation. Starting from an environment map, we extract 9 images and use Prism~\cite{dirik2025prism} to estimate their intrinsics. The images further serve as input to DiffusionLight-Turbo~\cite{Chinchuthakun2025DiffusionLightTurbo} for environment-map estimation and to a \ac{VLM}~\cite{wang2025internvl3} to produce a text description (see \cref{sec:Dataset} for details). 
    }
    \label{fig:dataset}
\end{figure}

\section{Dataset}
\label{sec:Dataset}

We aim to develop a joint embedding space that represents illumination across several modalities: $360\degree$ environment maps, regular images, irradiance maps, and text descriptions. In this section, we describe how we construct our dataset that encompasses and aligns these modalities. 

Specifically, we construct a new dataset to train and evaluate our multi-modal encoders, derived from a set of 8,020 \ac{HDR} environment maps combining both outdoor (ULaval Outdoor~\cite{hold2019deep}, Polyhaven~\cite{polyhaven2023poly}, Domeble~\cite{domeble2025}) and indoor sources (ULaval Indoor~\cite{Gardner_2017_TOG}, Polyhaven~\cite{polyhaven2023poly}, and pictures we acquired ourselves using a Ricoh Theta Z1 camera). We take 9 crops per environment map to generate a total of 72,180 multi-modal samples, each consisting of: (1)~the environment map itself, (2)~an image extracted from it, (3)~an irradiance representation of the image, and (4)~a text description for the image. \Cref{fig:dataset} shows an overview of our data generation pipeline; next we describe how we derive each aforementioned modality.

\paragraph{Image.}

We extract perspective images from an environment map by rotating the map about the vertical axis in $40\degree$ intervals and projecting it onto an image plane. Each extracted image has a resolution of $512\!\times\!512$ pixels and a $90\degree$ field of view.
We auto-expose the extracted image with Reinhard's luminance mapping \cite{reinhard2023photographic} with a scale $F_d\!=\!0.35$, and tone-map the result using $\gamma=2.2$. 


\paragraph{Irradiance.}

For each image, we compute the corresponding irradiance map using Prism~\cite{dirik2025prism}, a recent intrinsic decomposition method that can estimate spatially varying illumination. We include this representation as it is common in generative rendering pipelines~\cite{zeng2024rgbx, dirik2025prism}. 

\paragraph{Text.}

We generate natural-language descriptions of the lighting in each image using the InternVL3-38B \ac{VLM}~\cite{wang2025internvl3}. The bright light sources in the corresponding \ac{HDR} environment map are identified by detecting large connected components above an intensity threshold \cite{Gardner_2019_ICCV}. The threshold is initialized at a high value ($\tau_0=4$) and lowered by one stop ($\tau_{i+1}=\tau_i/\sqrt{2}$) until pixels above that value are found. The brightest point in each region above $\tau$ is kept as the dominant light source. The dominant positions are then integrated into a structured prompt for the \ac{VLM} to describe the illumination of the images extracted from the environment map, emphasizing directional cues. The \ac{VLM} is then queried with this prompt, the Reinhard-tone-mapped~\cite{reinhard2023photographic} environment map, and the extracted image. This procedure yields text descriptions that accurately capture key lighting directions, as illustrated in \cref{fig:dataset}.
The full prompt, our algorithm for brightest-region identification, and examples of text descriptions are detailed in the supplementary material.




\begin{figure*}[t]
    \centering
    \includegraphics{figures/overviewai_correcter.pdf}
    \vspace{-1mm}
    \caption{
        Overview of our embedding approach. Image- and text-based lighting modalities (see \cref{sec:Encoders}) are first embedded using DINOv2 and Qwen3, respectively. All modalities are then processed by lightweight fusion modules which are trained contrastively to output into our joint latent space, \method. To improve latent-space coherence, a prediction head estimates spherical-harmonics (SH) coefficients from the latents, and a dedicated loss aligns these coefficients to ground-truth coefficients extracted from the environment map.
    }
    \label{fig:light_encoders}
\end{figure*}

\paragraph{Spherical harmonics.}

To better capture directional lighting information, our model also estimates spherical harmonics (SH) coefficients, as described in \cref{sec:architecture}. We fit SH coefficients up to degree 3 to each environment map, to obtain a compact representation of the scene illumination.


\paragraph{Estimated environment map.}

To complement the high-quality environment maps in our dataset, we also estimate an environment map for each extracted image using DiffusionLight-Turbo~\cite{Chinchuthakun2025DiffusionLightTurbo}. These estimated maps are included during training to ensure encoder compatibility with them, as they can be easily generated from an image.

\section{Multi-modal embedding}
\label{sec:Method}

We propose a multi-modal embedding model that maps diverse lighting-related modalities - environment maps, irradiance maps, images, and text descriptions - into a shared latent space. Separate encoders are fine-tuned for image-based modalities and text, and their outputs are aligned using a lightweight transformer-based fusion module trained with a contrastive objective. The resulting latent space should place aligned lighting modalities, \eg corresponding environment maps and text descriptions, close to each other. We show an overview of our method in \cref{fig:light_encoders}.



\subsection{Encoders}
\label{sec:Encoders}

\paragraph{Image-based encoders.}

We employ a separate DINOv2-B~\cite{oquab2023dinov2} backbone for each image modality: environment map, image, and irradiance map.

To represent \ac{HDR} information from an environment map $I_{\text{hdr}}$, we convert it into both a \ac{LDR} image $I_{\text{ldr}}$ using Reinhard~\cite{reinhard2023photographic} tone-mapping, and a logarithmic encoding $I_{\text{log}} = \log({I_{\text{hdr}}+1}) / \log({I_{\text{max}}})$ following prior work~\cite{jin2024neural_gaffer, DiffusionRenderer}. We set $I_{\text{max}} = 1000$ as a normalization constant defining the upper bound of \ac{HDR} values, above which values are clipped. In addition, we provide a per-pixel x,y,z coordinate encoding $I_{\text{dir}}$ that corresponds to the latitude--longitude projection of the environment map, supplying explicit directional information. The complete encoder input for environment maps is thus $\{I_{\text{ldr}}, I_{\text{log}}, I_{\text{dir}}\}$ combining \ac{LDR}, \ac{HDR}, and directional cues. During training, environment maps are randomly sampled from either ground-truth data or DiffusionLight-Turbo~\cite{Chinchuthakun2025DiffusionLightTurbo} estimates to improve robustness to different input formats. We also apply random dropout to the $I_{\text{log}}$ channels to ensure compatibility with purely \ac{LDR} inputs.

For the image and irradiance-map input modalities, the DINOv2-B encoders are simply given their respective images as input with no further change.

\paragraph{Text encoder.}

We encode textual lighting descriptions using the 0.6B-parameter Qwen3 Embedding~\cite{qwen3embedding}, a large language model that produces general-purpose semantic embeddings from natural language. More specifically, we prompt the model with the lighting description and add additional instructions to encode the scene lighting, dominant light positions and overall brightness and color temperature. We include the exact prompt formatting in the supplemental material. We fine-tune this encoder within our multi-modal framework to better capture lighting-related semantics.

\subsection{Architecture}

\label{sec:architecture}
The modality-specific backbones (DINOv2 and Qwen3) produce features with varying sequence lengths $T_\text{backbone}$ and dimensions $D_\text{backbone}$. To unify these outputs, we introduce a summary module comprising $T$ learnable query tokens~\cite{li2023blip,pan2025transfer} per modality that attend to the backbone features through a multi-head attention layer followed by layer normalization. A linear projection then maps the resulting token features to a shared latent embedding $E \in \mathbb{R}^{T \times D}$, where we set $T=8$ and $D=512$ unless otherwise stated. We ablate the impact of the number of tokens in \cref{sec:model_eval}.\\
To encourage the latent space to capture directional lighting cues, we add a spherical-harmonics (SH) prediction head which takes our embedding tokens as input. The head estimates SH coefficients up to degree $l=3$, providing explicit supervision on lighting directionality during training.

\paragraph{Loss function.}

To force \method to closely align the latent codes for different modalities describing the same lighting condition, we employ a contrastive objective across all lighting modalities. Given a batch of samples, each containing an image, irradiance map, environment map, and lighting description, each modality is passed through its corresponding encoder (DINOv2 or Qwen3). We then compute pairwise cosine similarities between all modality embeddings and apply a cross-entropy contrastive loss $\mathcal{L}_{\text{C}}$ to maximize agreement between matching pairs. To enforce directional consistency, we add an SH supervision term based on the mean squared error between the predicted and ground-truth coefficients: $\mathcal{L}_{\text{SH}} = \lVert \text{SH}_{\text{pred}} - \text{SH}_{\text{GT}} \rVert_2^2$. The total loss combines both objectives: $\mathcal{L} = \mathcal{L}_{\text{C}} + \mathcal{L}_{\text{SH}}$.







\section{Experiments}

\begin{figure}[t]
    \centering
    \includegraphics[width=\linewidth]{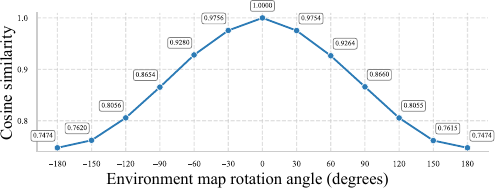}
    \vspace{-6mm}
    \caption{
        Analysis of light direction encoding in our unified representation. The environment map is rotated about the vertical axis (x-axis), and the resulting cosine similarity against the original orientation is shown. Similarity decreases with increasing rotation, indicating that the latent features explicitly encode light direction.
    }
    \label{fig:envmap_experiments}
\end{figure}

We evaluate our model and demonstrate our joint latent embedding on three applications: cross-modal retrieval, environment-map generation, and image relighting. We show that our \method representation enables both light understanding and editing with diffusion models~\cite{SD3}.

\subsection{Model evaluation}
\label{sec:model_eval}

Our approach encodes multiple lighting modalities into a joint latent space. We evaluate the sensitivity of our latent embeddings to lighting direction and analyze our design choices through ablation studies.

\paragraph{Light direction.}

We rotate environment maps from our test set horizontally from $-180\degree$ to $+180\degree$ in $30\degree$ increments and compute cosine similarity between transformed and initial (i.e., $0\degree$) embeddings. \Cref{fig:envmap_experiments} shows that similarity decreases with larger rotations, indicating our embeddings successfully capture directional lighting information.

\begin{figure}[t]
    \centering
    \footnotesize
    \setlength{\tabcolsep}{1pt}
    \includegraphics[width=\linewidth]{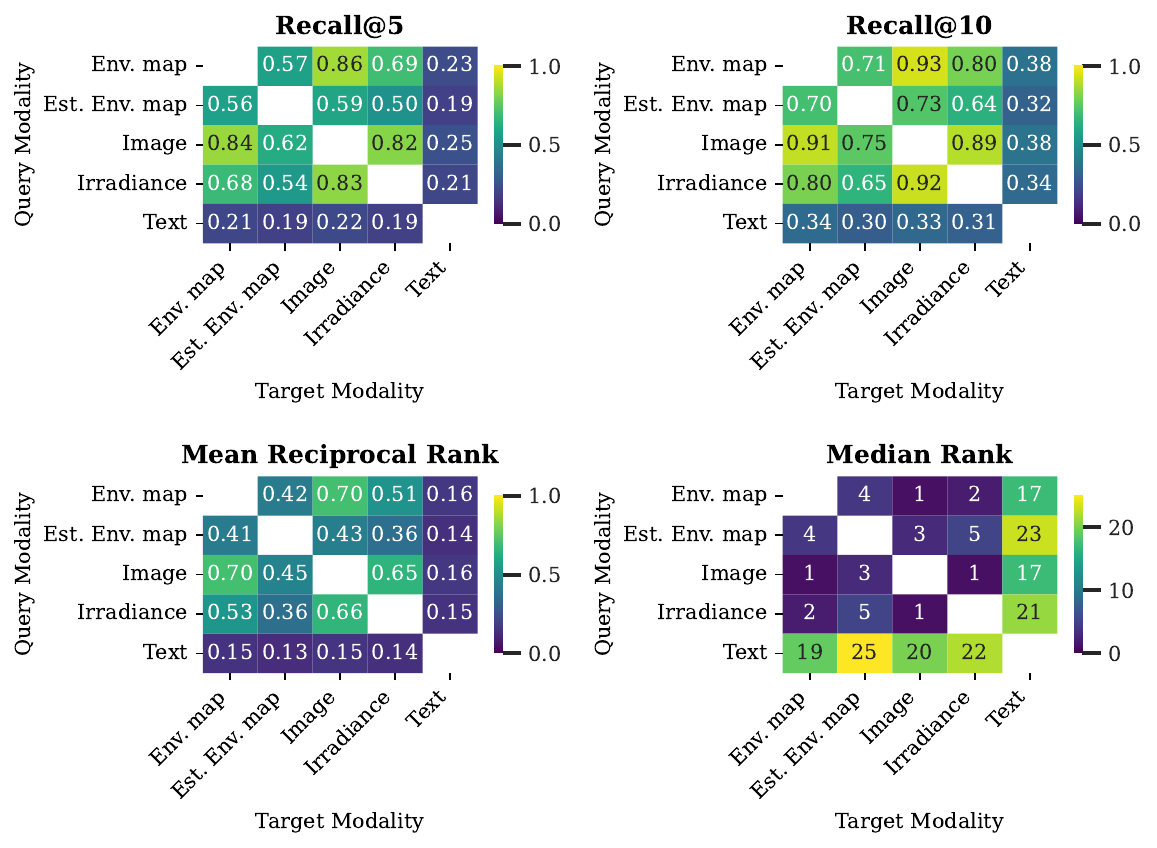}
    \vspace{-6mm}
    \caption{
        Recall rate (top, higher is better), mean reciprocal rank (bottom left, higher is better), and median rank (bottom right, lower is better) between different modalities.
    }
    \label{fig:retrieval_stats}
\end{figure}

\begin{table}[t]
    \centering
    \caption{
        Ablation study. Retrieval metrics for varying numbers of tokens or SH degrees, computed over 603 test-data points. We report averaged retrieval statistics for Image$\to$Text and Text$\to$Image on CLIP embeddings for comparison.
    }
    \label{tab:ablation_retrieval_metrics}
    \scriptsize
    \setlength{\tabcolsep}{3.8pt}
    \begin{tabular}{l|cccccc}
        \toprule
        \textbf{Variant} & \textbf{R@1} & \textbf{R@5} & \textbf{R@10} & \textbf{MRR} & \textbf{Median rank} & \textbf{Mean rank} \\
        \midrule
        CLIP VIT-B/32 & 2.6 & 10.8 & 16.9 & 0.077 & 72.0 & 107.3 \\
        Qwen3-VL 2B & 8.9 & 26.3 & 37.1 & 0.179 & 36.6 & 56.4 \\
        \midrule
        1 tokens, SH3 & 23.8 & 47.4 & 58.9 & 0.355 & 10.7 & 22.2 \\
        2 tokens, SH3 & 24.3 & 48.7 & 60.1 & 0.362 & 9.9 & 21.2 \\
        4 tokens, SH3 & 25.1 & 49.0 & 60.5 & 0.369 & 10.1 & 21.5 \\
        8 tokens, SH3 & 24.9 & 49.0 & 60.6 & 0.367 & 9.8 & 21.2 \\
        16 tokens, SH3 & \textbf{26.5} & \textbf{50.7} & \textbf{61.4} & \textbf{0.382} & \textbf{9.6} & \textbf{21.2} \\ \midrule
        8 tokens, NOSH & 10.2 & 31.9 & 45.0 & 0.215 & 15.5 & 29.7 \\
        8 tokens, SH1 & 20.7 & 43.3 & 54.4 & 0.320 & 14.2 & 25.9 \\
        8 tokens, SH3 & \textbf{24.9} & \textbf{49.0} & \textbf{60.6} & \textbf{0.367} & \textbf{9.8} & \textbf{21.2} \\
        8 tokens, SH5 & 24.1 & 47.7 & 59.5 & 0.358 & 10.2 & 21.5 \\
        \bottomrule
    \end{tabular}
\end{table}

\begin{figure*}[t]
    \footnotesize
\setlength{\tabcolsep}{3pt} 

\begin{tabular}{
    m{0.01\textwidth}
    m{0.18\textwidth}
    m{0.18\textwidth}
    m{0.18\textwidth}
    m{0.18\textwidth}
    m{0.18\textwidth}
}
    \rotatebox[origin=c]{90}{Inputs} &
    \includegraphics[width=0.18\textwidth]{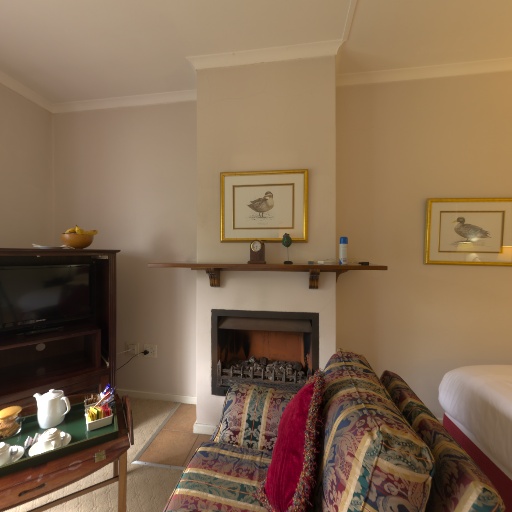} &
    \includegraphics[width=0.18\textwidth]{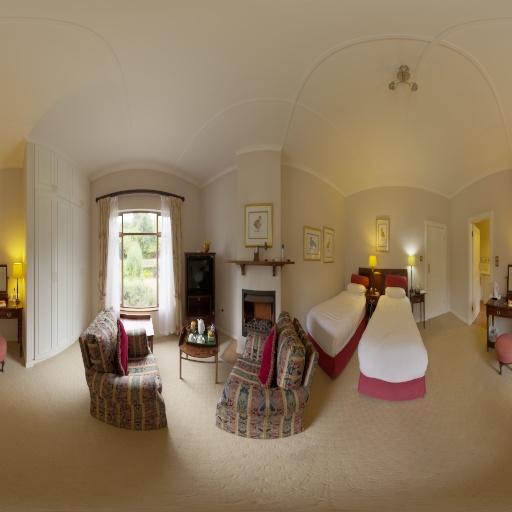} &
    \includegraphics[width=0.18\textwidth]{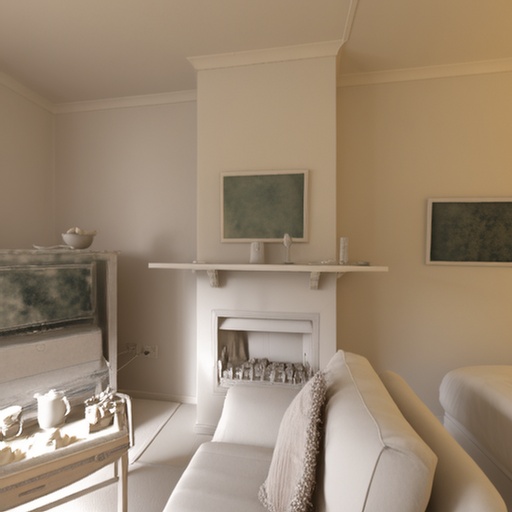} &
    \textit{``indoor, natural light from the left, warm glow, two table lamps with warm yellow light, balanced illumination, cozy ambiance.''} &  \\[-2pt]
    
    \vspace{-7mm}\rotatebox[origin=c]{90}{Rendered SH} &
    \includegraphics[width=0.18\textwidth]{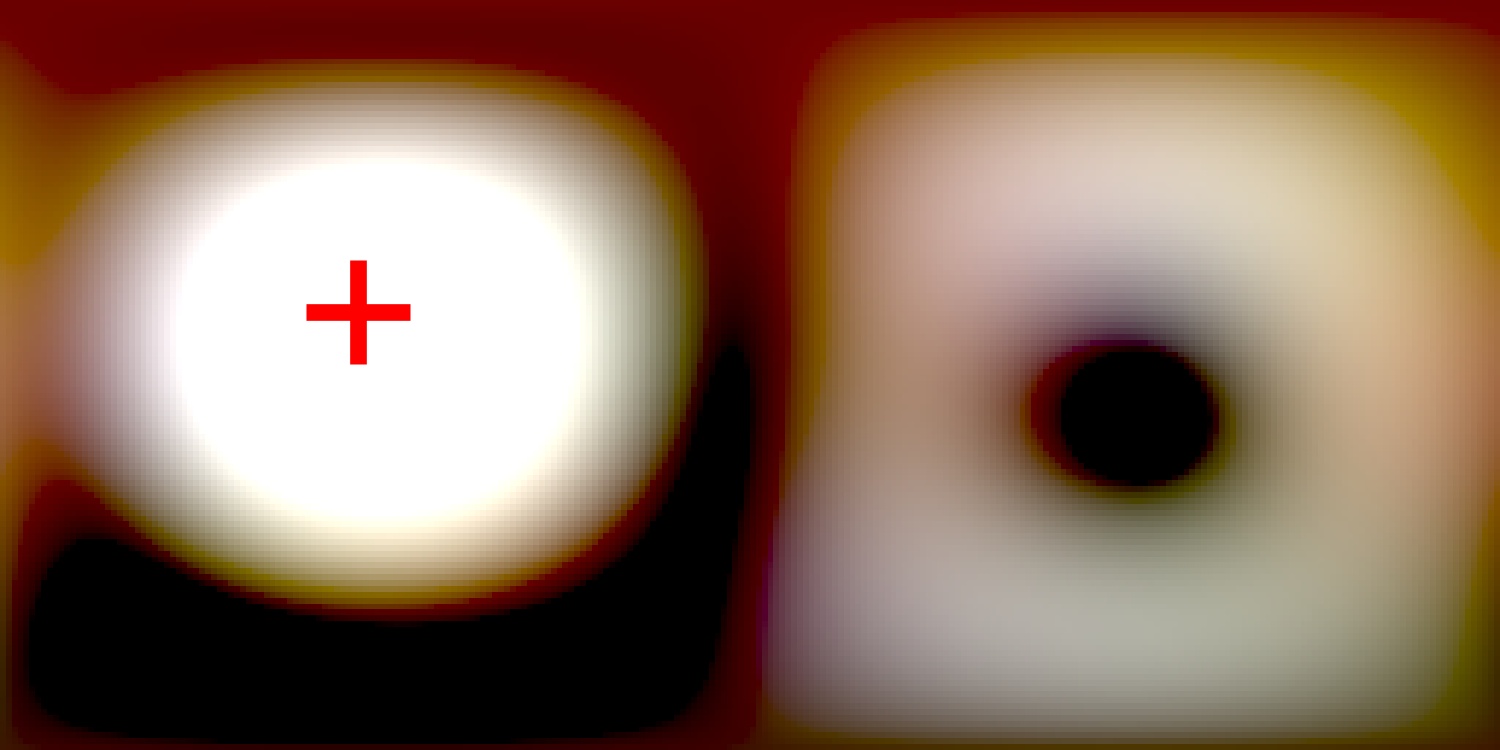} &
    \includegraphics[width=0.18\textwidth]{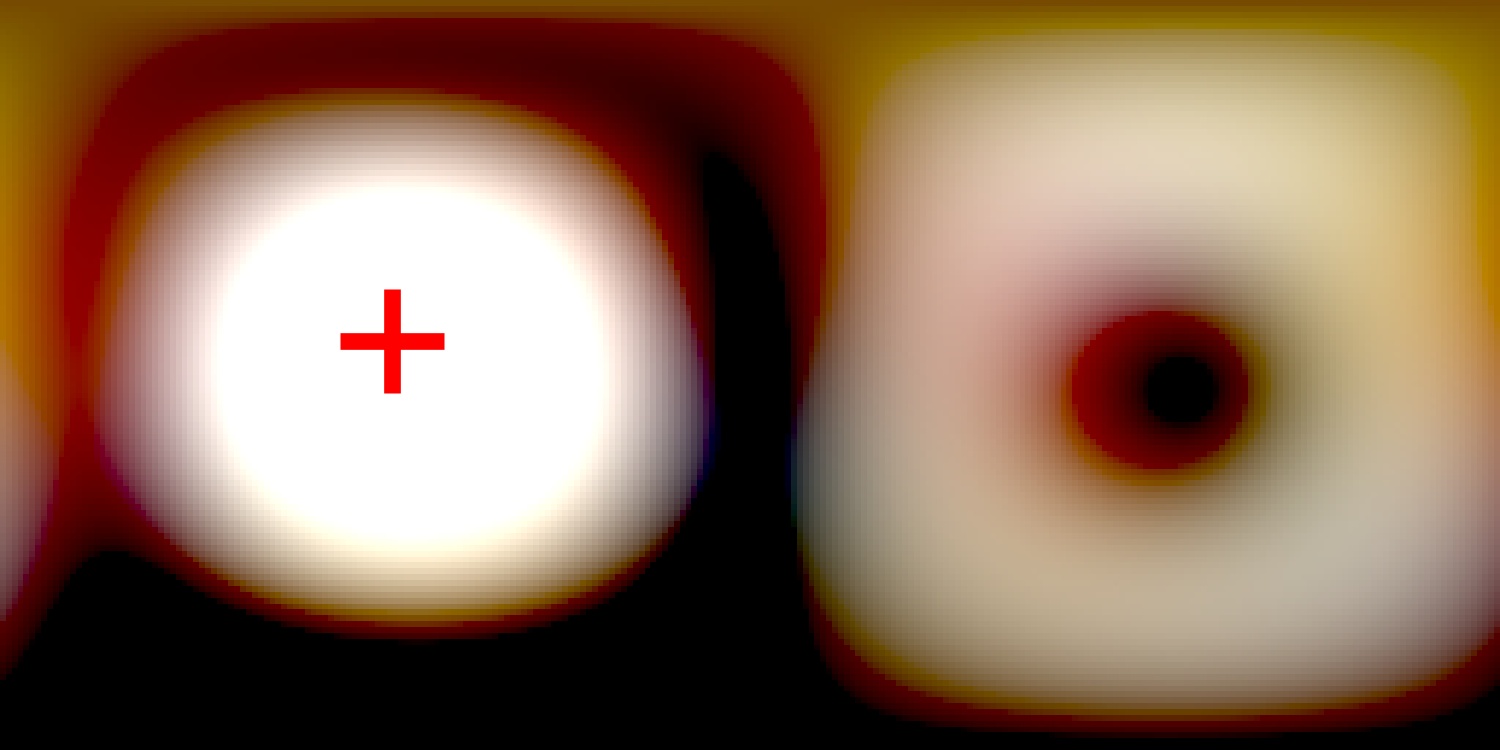} &
    \includegraphics[width=0.18\textwidth]{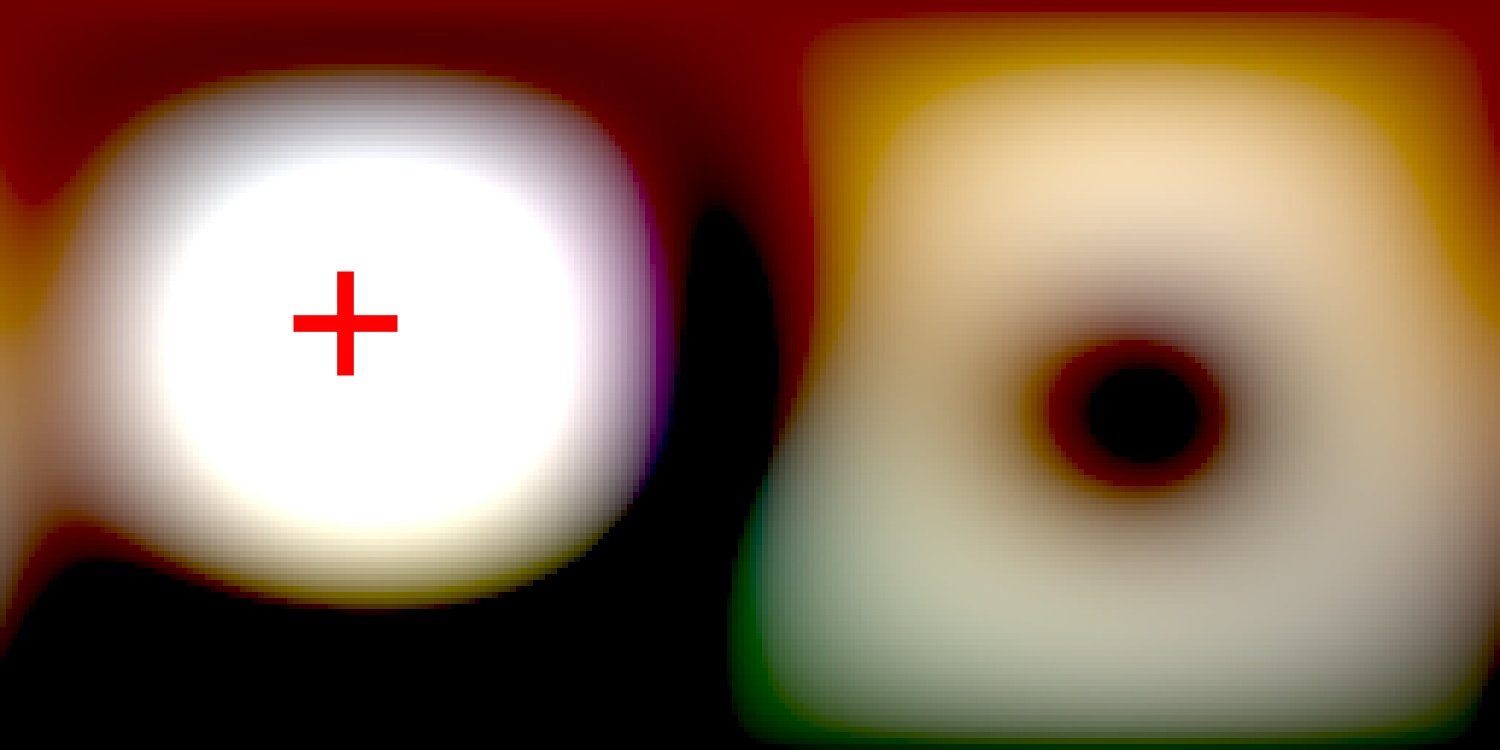} &
    \includegraphics[width=0.18\textwidth]{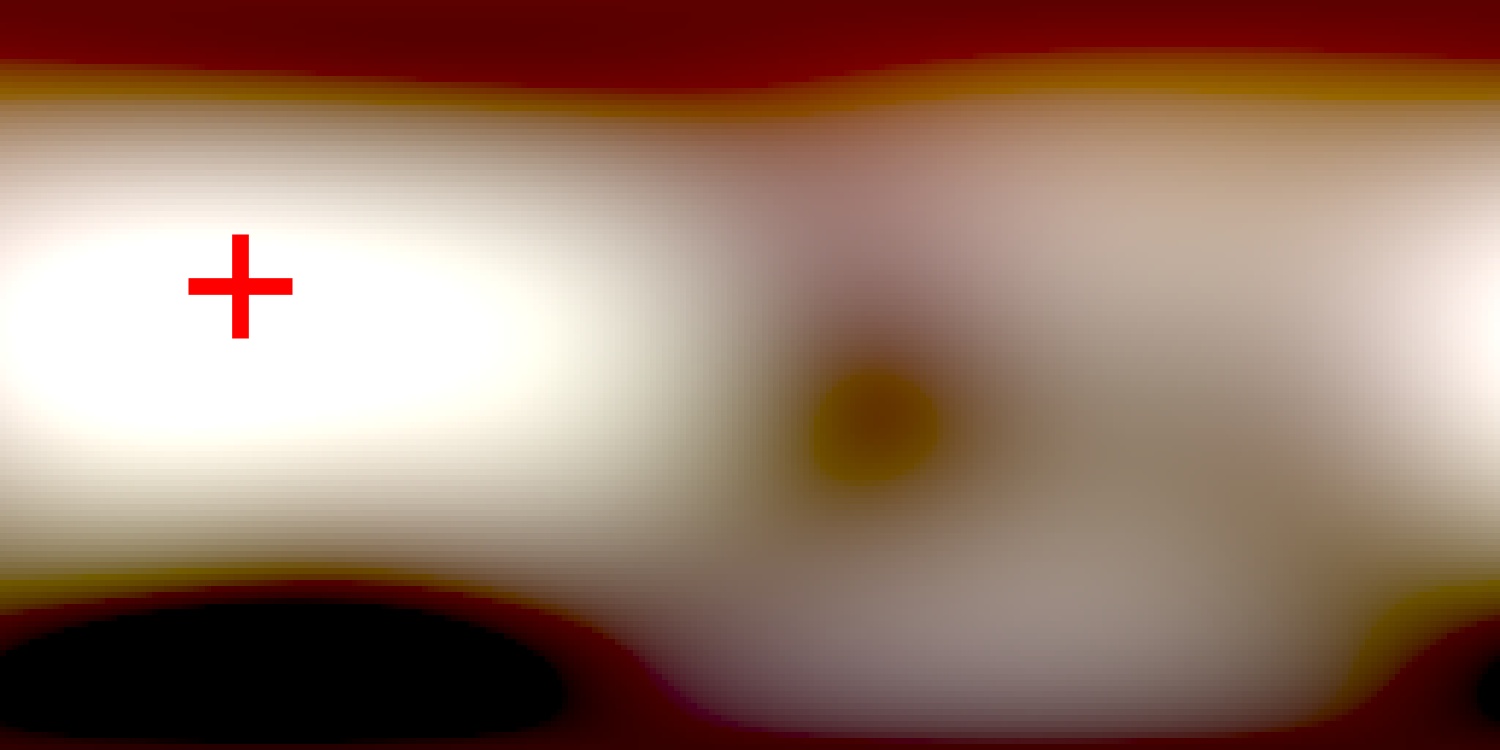} &
    \includegraphics[width=0.18\textwidth]{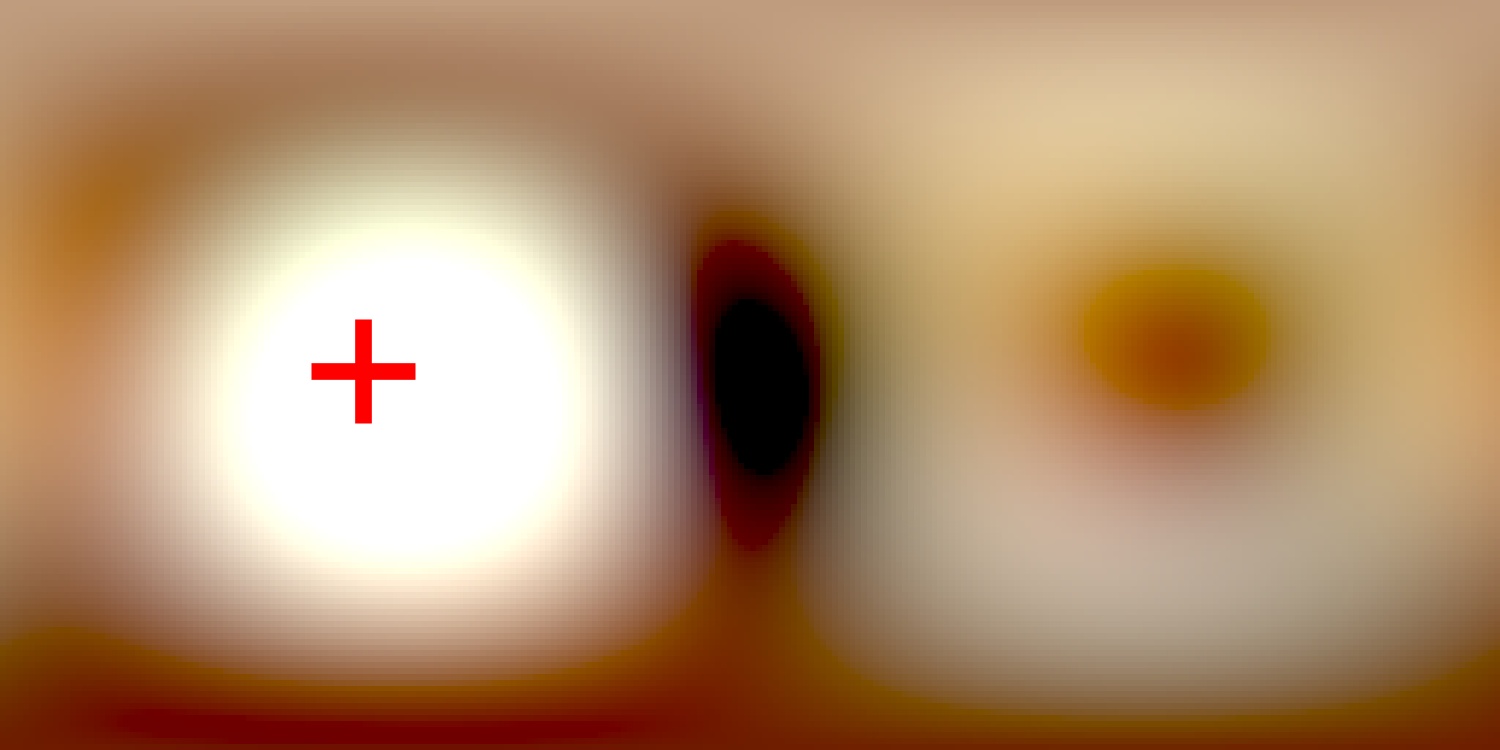} \\[10pt]
    
    \rotatebox[origin=c]{90}{Inputs} &
    \includegraphics[width=0.18\textwidth]{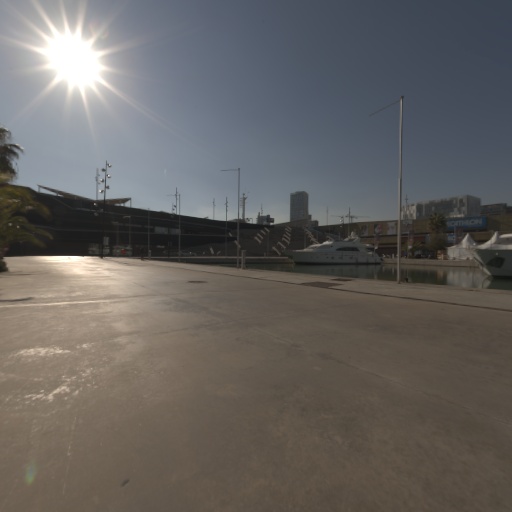} &
    \includegraphics[width=0.18\textwidth]{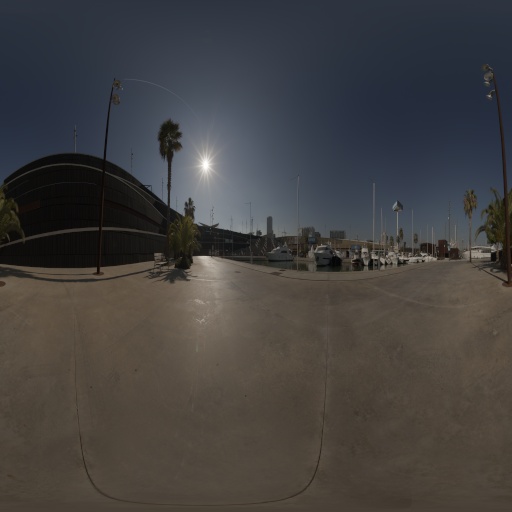} &
    \includegraphics[width=0.18\textwidth]{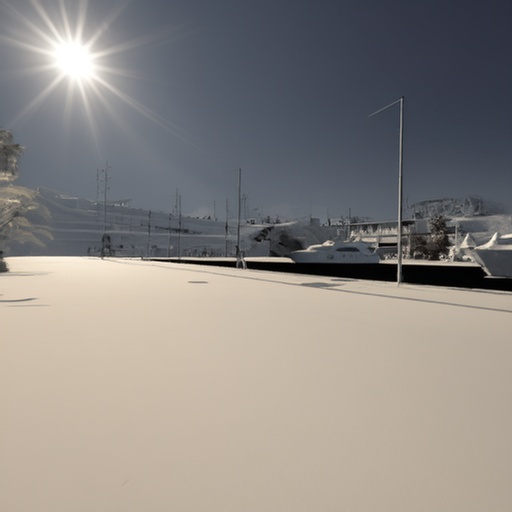} &
    \textit{``outdoor, bright sun from the upper left, warm golden hue, strong direct light, inactive streetlights, warm scene color, contrasting shadows''} & \\[-2pt]
    
    \vspace{-7mm}\rotatebox[origin=c]{90}{Rendered SH} &
    \includegraphics[width=0.18\textwidth]{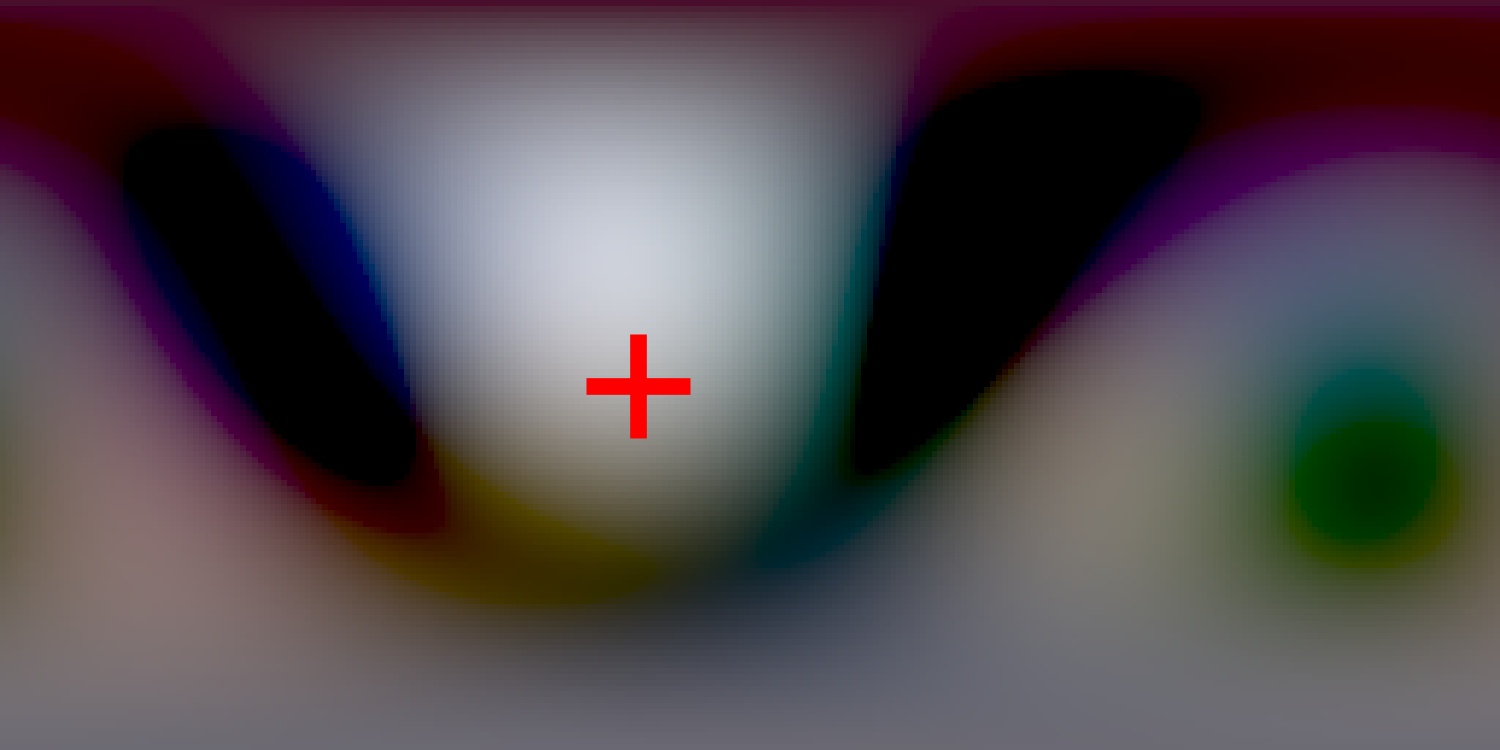} &
    \includegraphics[width=0.18\textwidth]{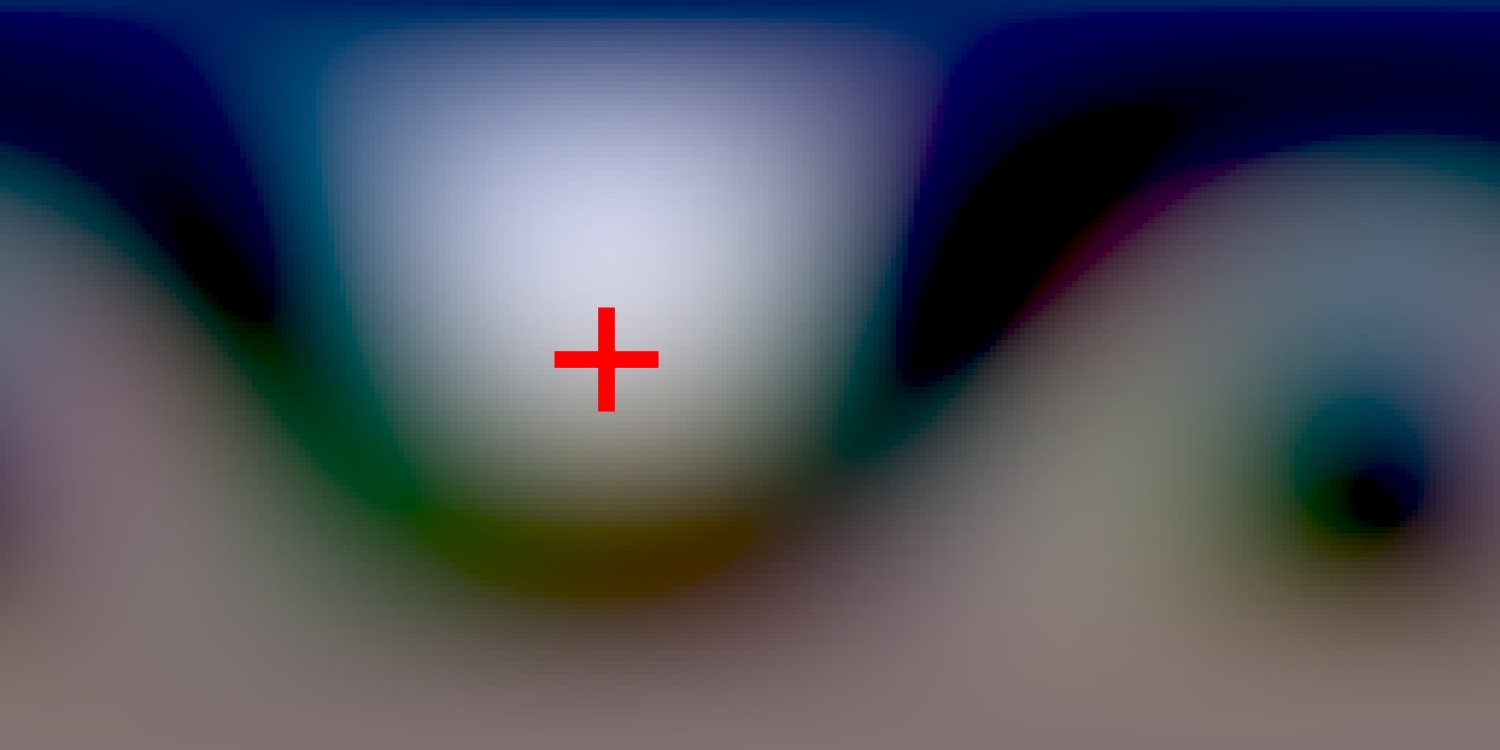} &
    \includegraphics[width=0.18\textwidth]{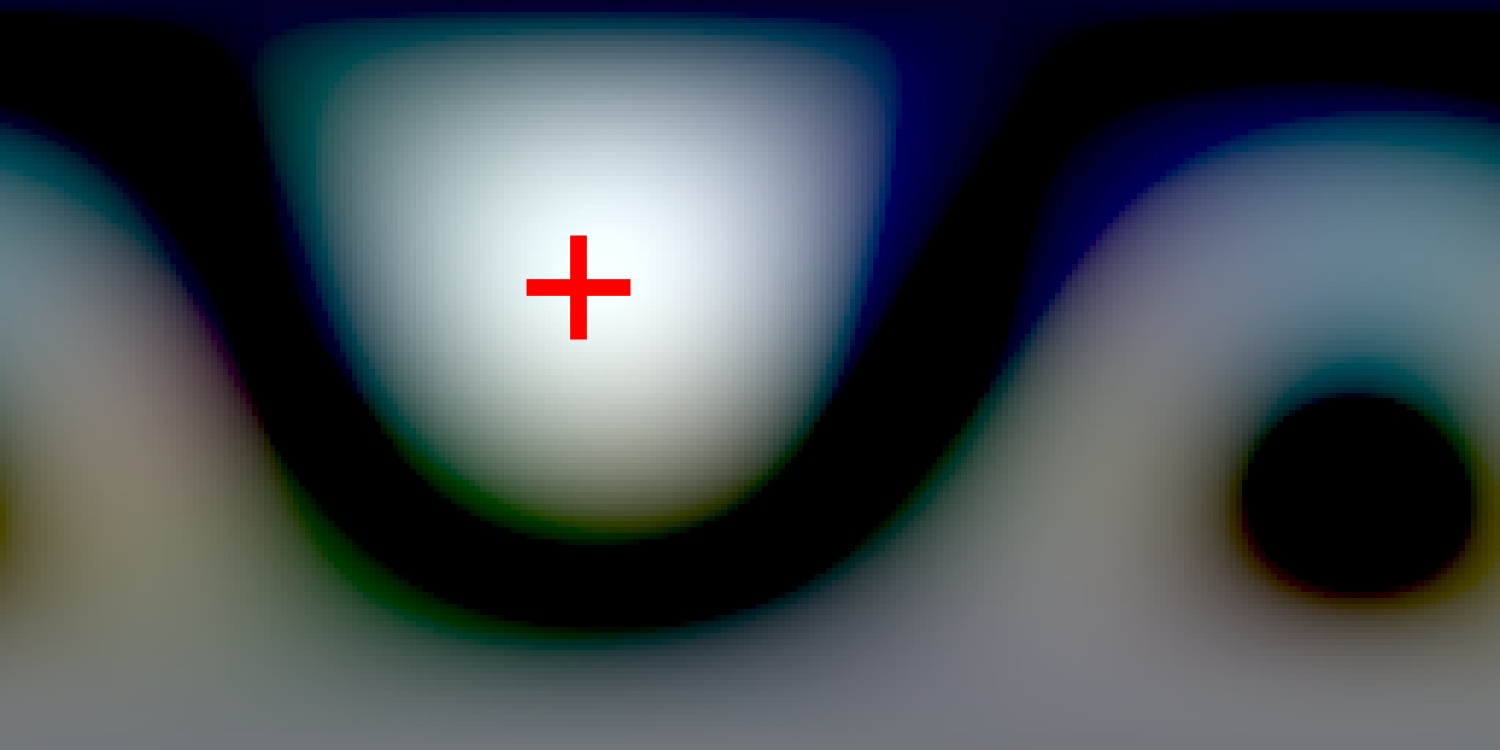} &
    \includegraphics[width=0.18\textwidth]{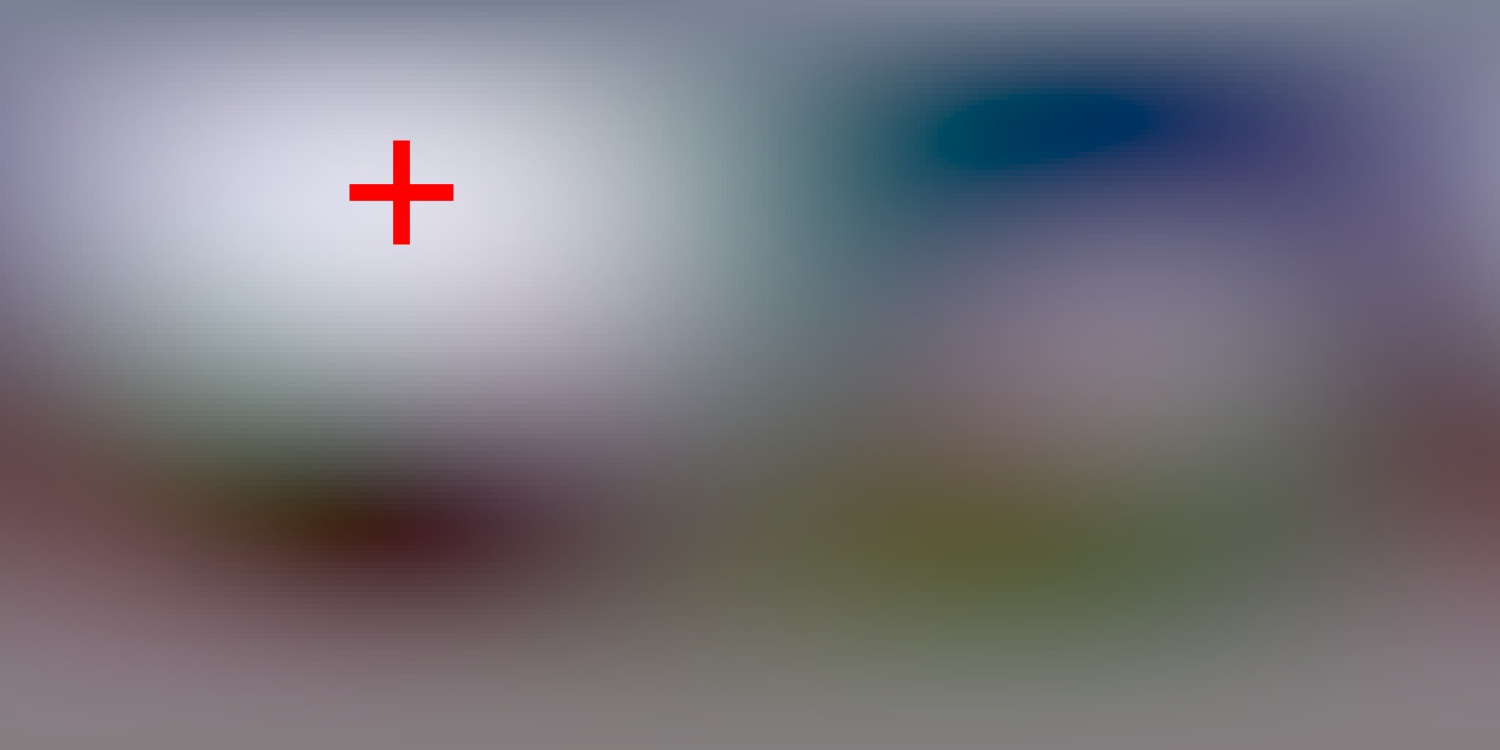} &
    \includegraphics[width=0.18\textwidth]{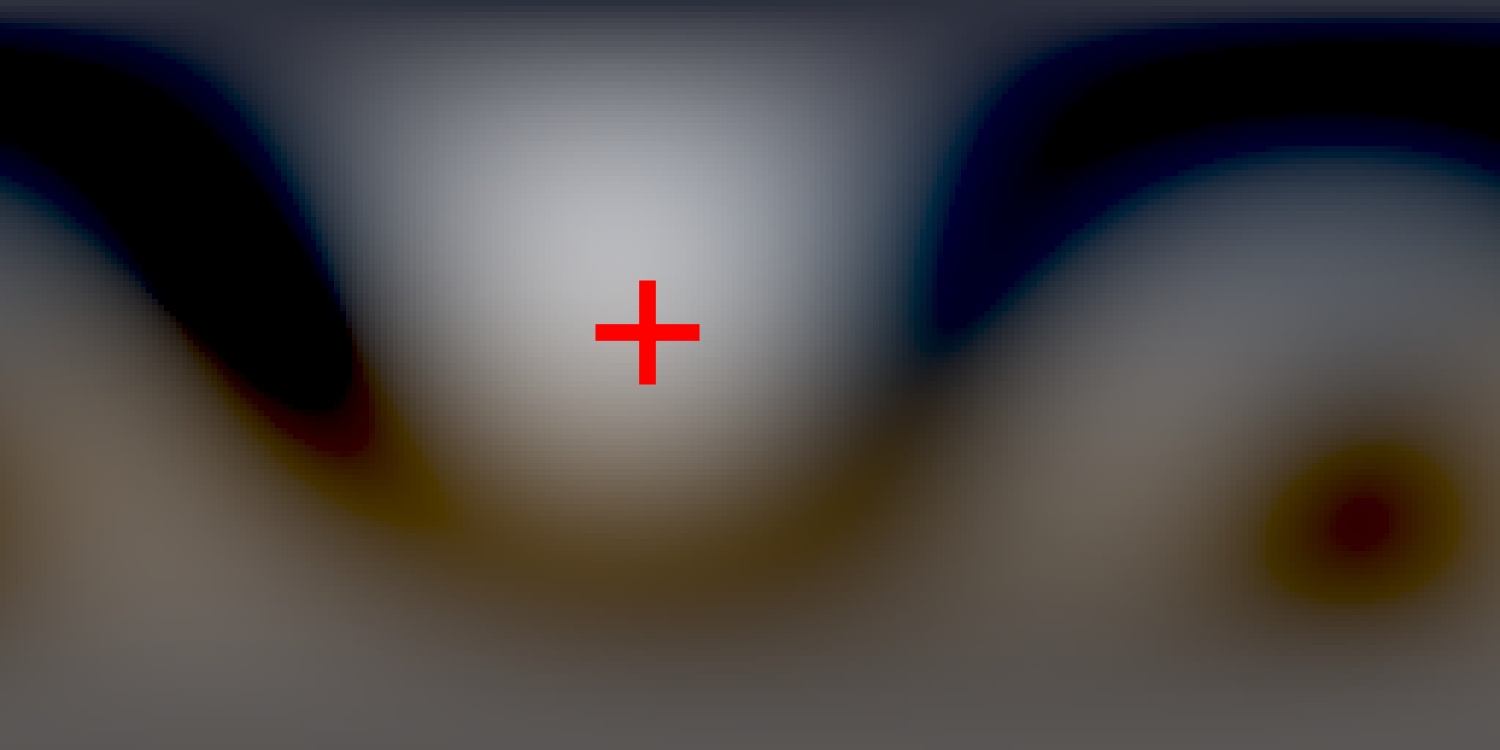} \\[6pt]
    
    & \centering Image & \centering Environment map & \centering Irradiance & \centering Text & \centering Ground truth \\
\end{tabular}
\vspace{-2mm}

    \caption{
        We visualize the SH coefficients extracted by our SH head (see \cref{fig:light_encoders}) from the \method embeddings of the input modalities (rows 1 and 3). We render the predicted SH to an environment map (rows 2 and 4) and visualize the dominant light direction as a red cross. Note how both the reconstructed environment maps and the dominant light directions align across modalities as well as with the reference, indicating that they indeed have similar latent embeddings.
    }
    \label{fig:sh_results}
\end{figure*}


\paragraph{SH reconstruction.}

We evaluate the light-encoding capabilities of our representation using our model's SH-prediction head. For a sample in our test dataset, we obtain embeddings for all modalities and render environment maps from the SH representations predicted from each embedding. In \cref{fig:sh_results} we compare these maps to a ground truth rendered from the SH representation extracted directly from the environment map in the dataset sample. We additionally mark the location of the dominant light direction in each map (given by the order-1 SH coefficients) with a red cross. We observe close alignment between all modalities and the ground truth, indicating that our embeddings accurately encode structural lighting information. Even though the text description carries much less information than other modalities, its SH representation exhibits good alignment.

\paragraph{Ablation study.}

We ablate our \method encoders on the number of embedding tokens (1, 4, 8, or 16), and spherical harmonics degree supervision by comparing retrieval statistics in \cref{tab:ablation_retrieval_metrics}. The retrieval is performed on our test set, containing 603 data points, across all modality pairs bidirectionally (e.g., environment map\,$\rightarrow$\,text and text\,$\rightarrow$\,environment map), and averaged. We can see that increasing the number of tokens offers small improvements in retrieval accuracy at the cost of dimensionality/memory. We chose 8 tokens, but different methods may prefer a tighter or larger representation to satisfy memory or precision constraints. In which case, 1 or 16 token(s) may respectively be preferred. We also show that spherical harmonics of degree 3 (SH3) yield the greatest improvement in retrieval, while removing the spherical harmonics direction supervision (NOSH) significantly degrades retrieval accuracy. 

\subsection{Retrieval}

We evaluate our shared latent space through cross-modal retrieval. Given a query embedding obtained from an input lighting modality, we aim to retrieve other modalities with similar lighting. We first extract embeddings from all modalities in our test set, consisting of 603 entries. We then compute pairwise cosine similarities between the query embedding and all other embeddings. 

\begin{figure*}
    \centering
\setlength{\tabcolsep}{1pt}    
\renewcommand{\arraystretch}{0.5} 

\begin{tabular}{
    m{0.01\textwidth}
    >{\centering\arraybackslash}m{0.13\textwidth}
    m{0.13\textwidth}
    m{0.13\textwidth}
    m{0.13\textwidth}
    @{\hspace{0.025\textwidth}} 
    m{0.13\textwidth}
    m{0.13\textwidth}
    m{0.13\textwidth}
}



\vspace{-3.0cm}\rotatebox[origin=c]{90}{\scriptsize{Image $\to$ Env.\,maps}} & \begin{overpic}[width=\linewidth]{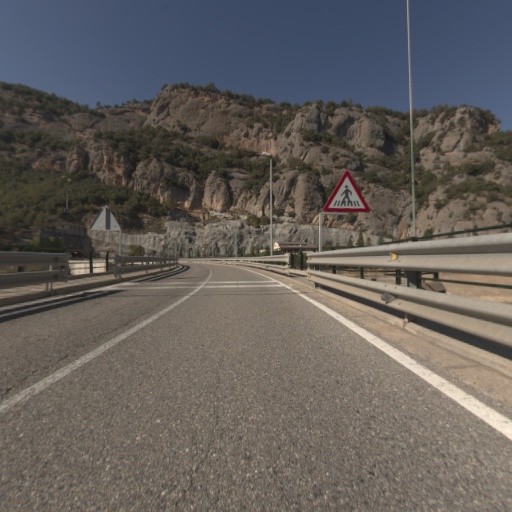}\end{overpic} &
\begin{overpic}[width=\linewidth]{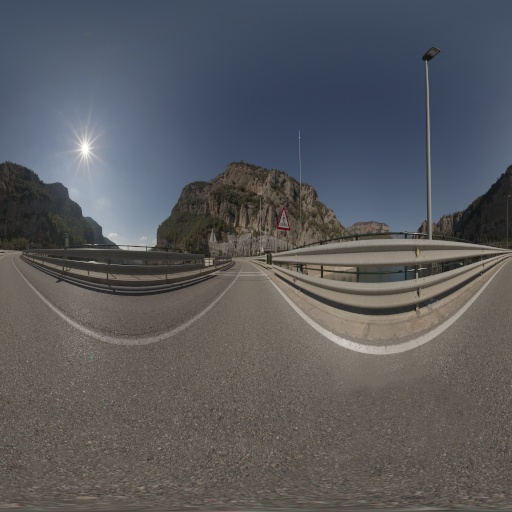}\put(6,10){\color{white}\bfseries \scriptsize 0.996}\end{overpic} &
\begin{overpic}[width=\linewidth]{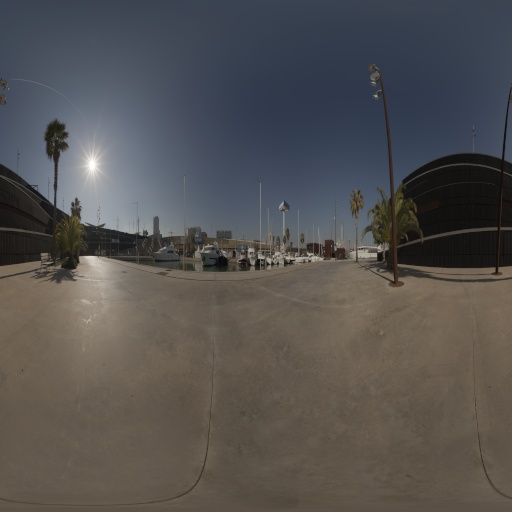}\put(6,10){\color{white}\bfseries \scriptsize 0.996}\end{overpic} &
\begin{overpic}[width=\linewidth]{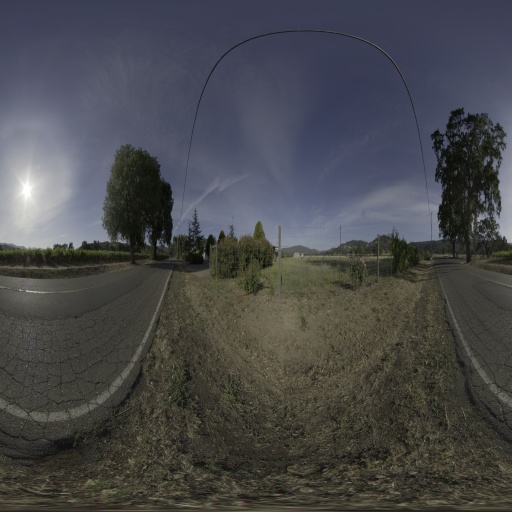}\put(6,10){\color{white}\bfseries\scriptsize  0.960}\end{overpic} &
\begin{overpic}[width=\linewidth]{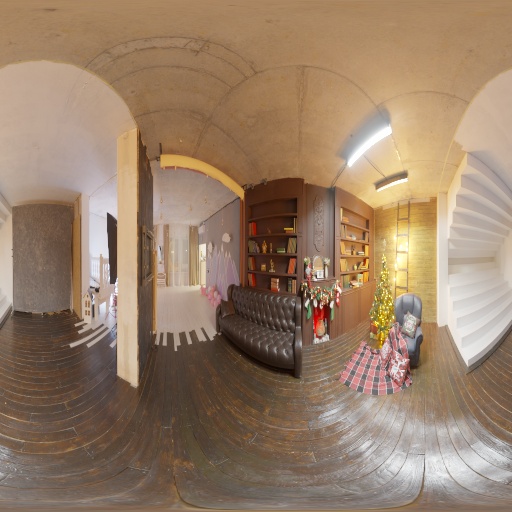}\put(6,10){\color{white}\bfseries\scriptsize  -0.562}\end{overpic} &
\begin{overpic}[width=\linewidth]{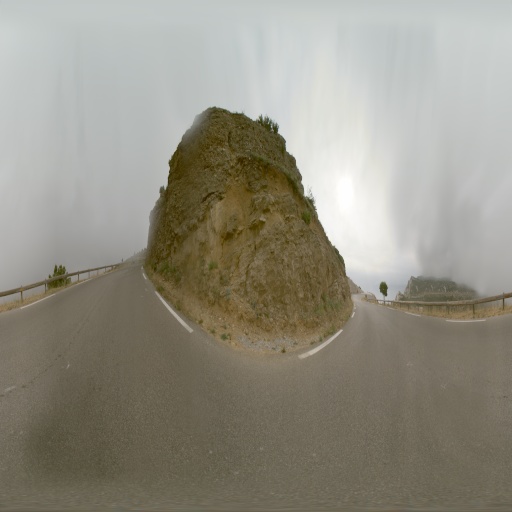}\put(6,10){\color{white}\bfseries\scriptsize  -0.535}\end{overpic} &
\begin{overpic}[width=\linewidth]{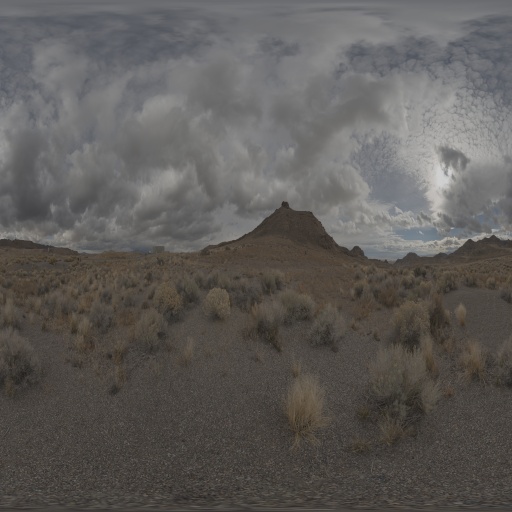}\put(6,10){\color{white}\bfseries\scriptsize  -0.511}\end{overpic} \\[2pt]

\vspace{-1.2cm}\rotatebox[origin=c]{90}{\scriptsize{Text $\to$ Env.\,maps}} & \begin{minipage}[c]{\linewidth}\centering\vspace{-5mm}\scriptsize{\textit{``indoor, natural light from the left, warm glow, two table lamps with warm yellow light, balanced illumination, cozy ambiance.''}}\end{minipage} &
\begin{overpic}[width=\linewidth]{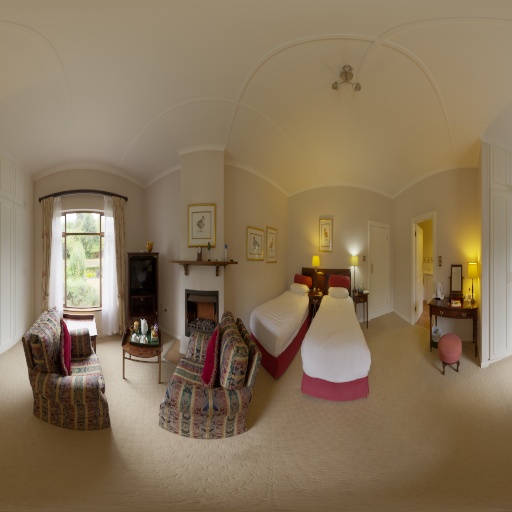}\put(6,10){\color{white}\bfseries\scriptsize  0.991}\end{overpic} &
\begin{overpic}[width=\linewidth]{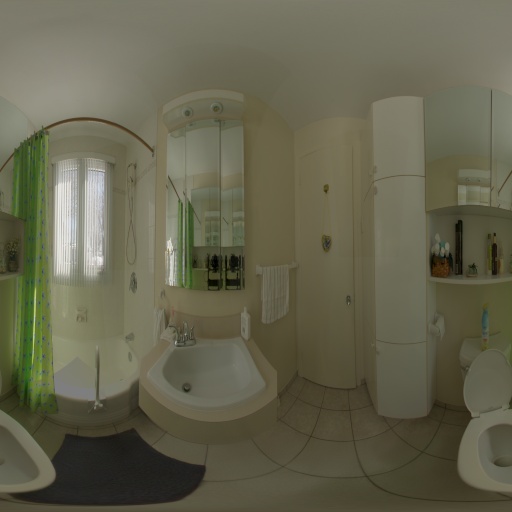}\put(6,10){\color{white}\bfseries\scriptsize  0.976}\end{overpic} &
\begin{overpic}[width=\linewidth]{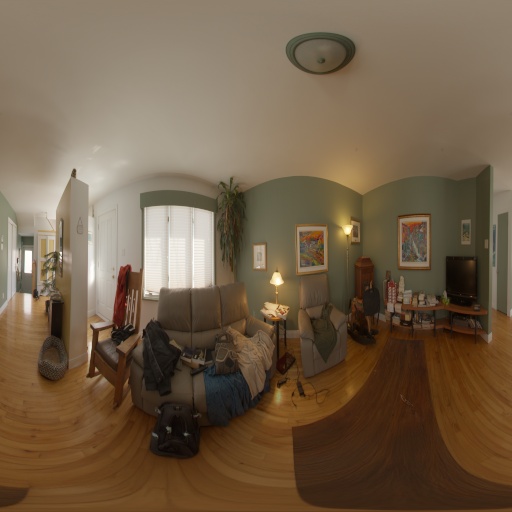}\put(6,10){\color{white}\bfseries\scriptsize  0.972}\end{overpic} &
\begin{overpic}[width=\linewidth]{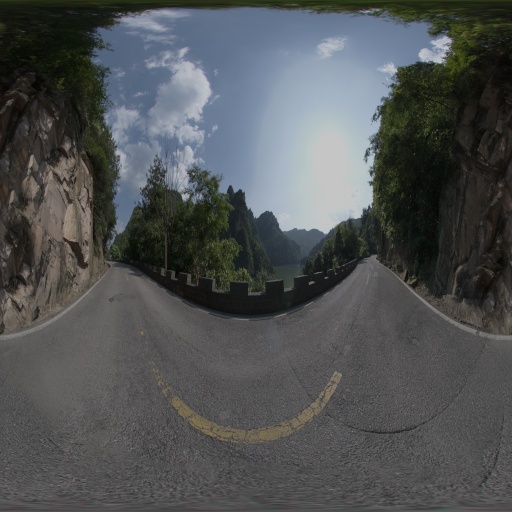}\put(6,10){\color{white}\bfseries\scriptsize  -0.584}\end{overpic} &
\begin{overpic}[width=\linewidth]{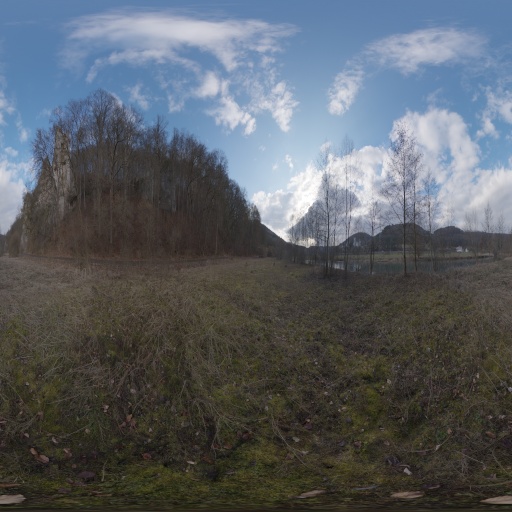}\put(6,10){\color{white}\bfseries\scriptsize  -0.543}\end{overpic} &
\begin{overpic}[width=\linewidth]{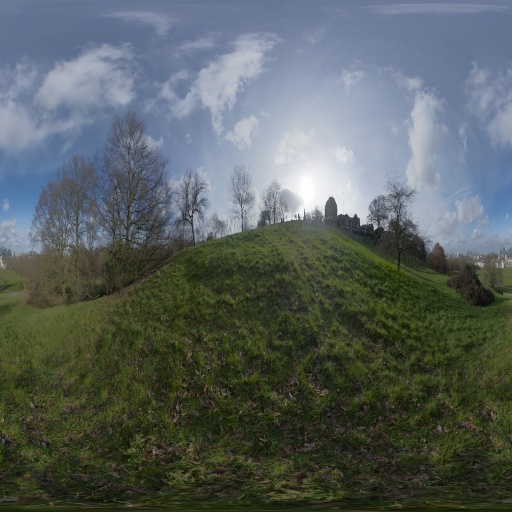}\put(6,10){\color{white}\bfseries\scriptsize  -0.541}\end{overpic} \\[2pt]


\vspace{-1.5cm}\rotatebox[origin=c]{90}{\scriptsize{Irradiance $\to$ Env.\,maps}} & \begin{overpic}[width=\linewidth]{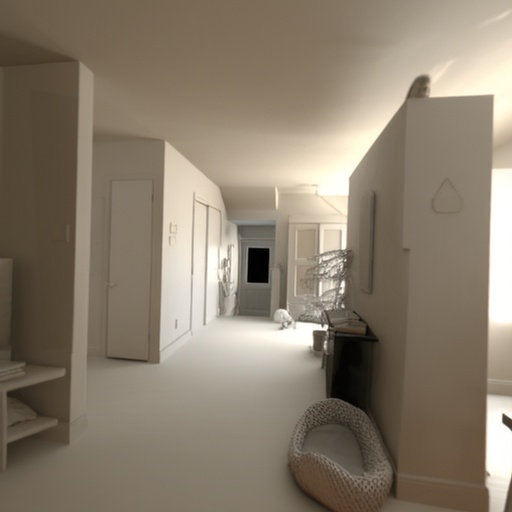}\end{overpic} &
\begin{overpic}[width=\linewidth]{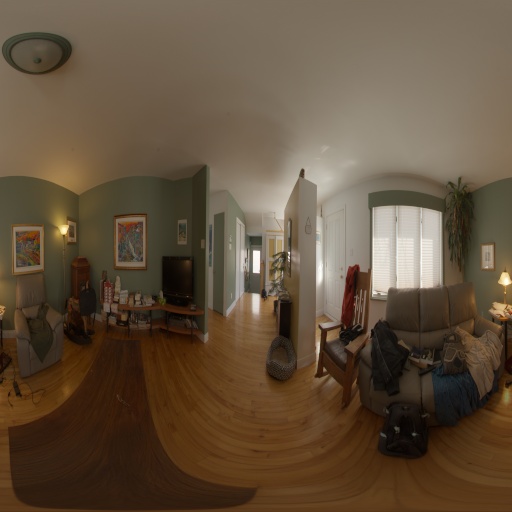}\put(6,10){\color{white}\bfseries\scriptsize  0.990}\end{overpic} &
\begin{overpic}[width=\linewidth]{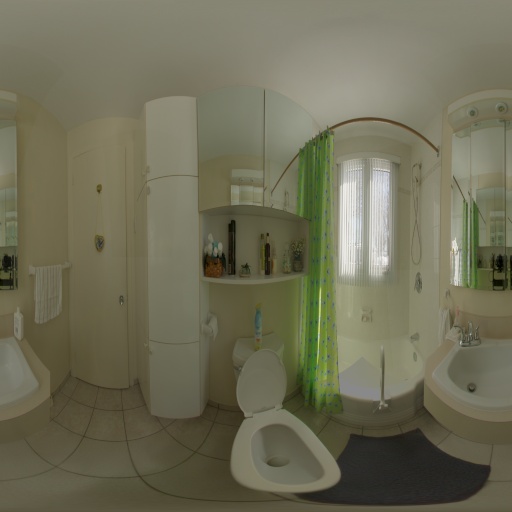}\put(6,10){\color{white}\bfseries\scriptsize  0.990}\end{overpic} &
\begin{overpic}[width=\linewidth]{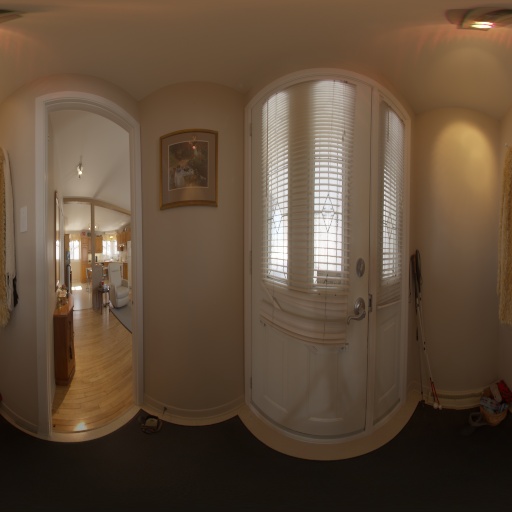}\put(6,10){\color{white}\bfseries\scriptsize  0.989}\end{overpic} &
\begin{overpic}[width=\linewidth]{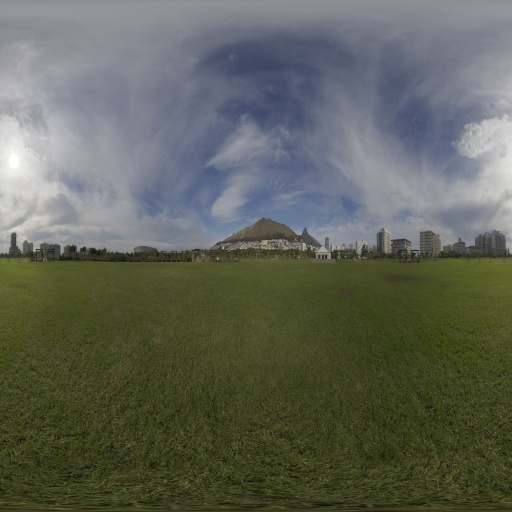}\put(6,10){\color{white}\bfseries\scriptsize  -0.521}\end{overpic} &
\begin{overpic}[width=\linewidth]{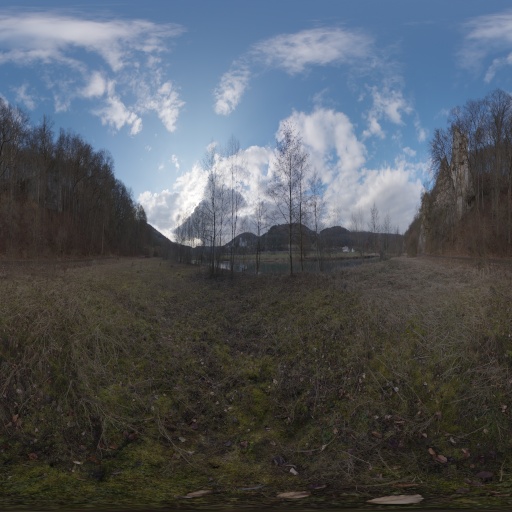}\put(6,10){\color{white}\bfseries\scriptsize  -0.521}\end{overpic} &
\begin{overpic}[width=\linewidth]{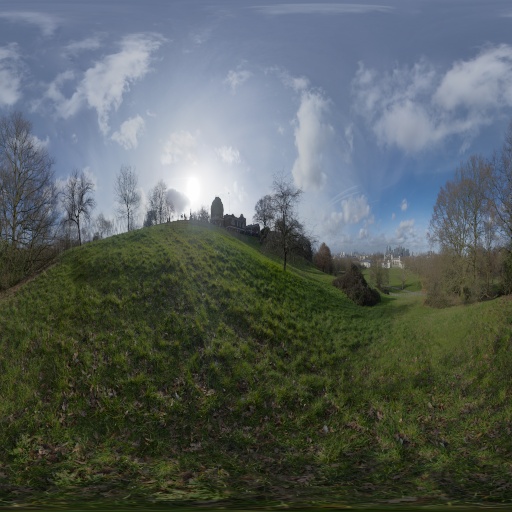}\put(6,10){\color{white}\bfseries\scriptsize  -0.516}\end{overpic} \\[2pt]

\vspace{-1.5cm}\rotatebox[origin=c]{90}{\scriptsize{Text $\to$ Image}} & \begin{minipage}[c]{\linewidth}\centering\vspace{-5mm}\scriptsize{\textit{``outdoor, bright sun from the upper right, warm golden hue, clear blue sky, reflections on water, no artificial lights.''}}\end{minipage} &
\begin{overpic}[width=\linewidth]{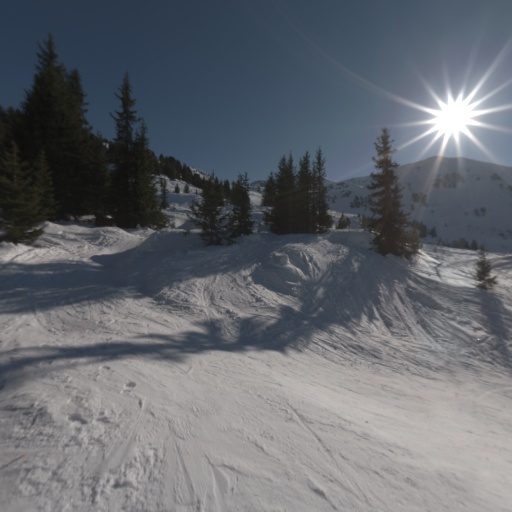}\put(6,10){\color{white}\bfseries\scriptsize  0.935}\end{overpic} &
\begin{overpic}[width=\linewidth]{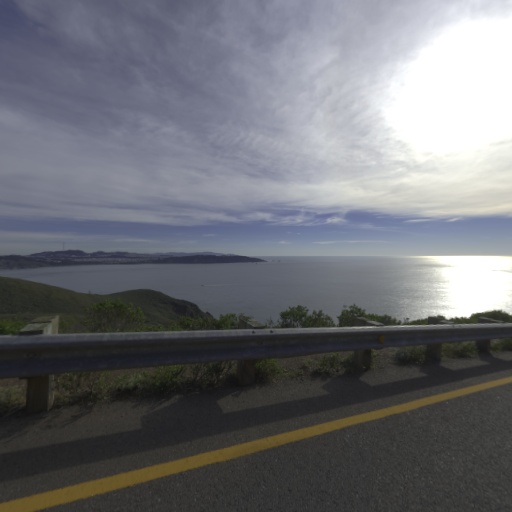}\put(6,10){\color{white}\bfseries\scriptsize  0.912}\end{overpic} &
\begin{overpic}[width=\linewidth]{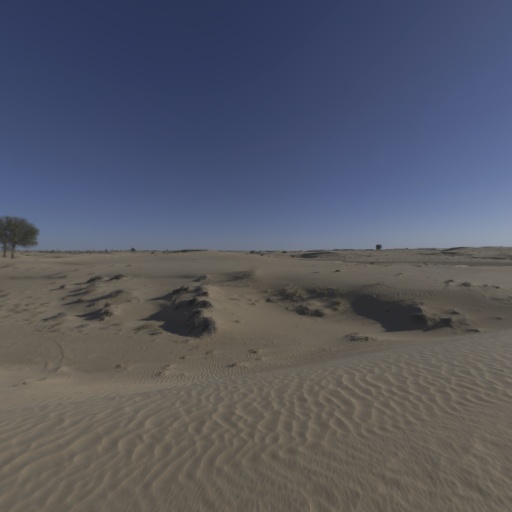}\put(6,10){\color{white}\bfseries\scriptsize  0.901}\end{overpic} &
\begin{overpic}[width=\linewidth]{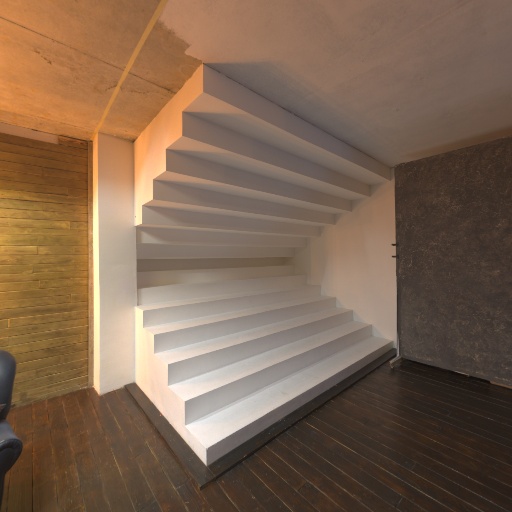}\put(6,10){\color{white}\bfseries\scriptsize  -0.621}\end{overpic} &
\begin{overpic}[width=\linewidth]{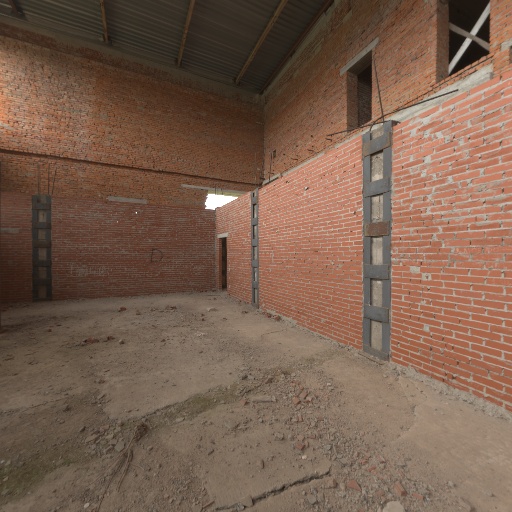}\put(6,10){\color{white}\bfseries\scriptsize  -0.609}\end{overpic} &
\begin{overpic}[width=\linewidth]{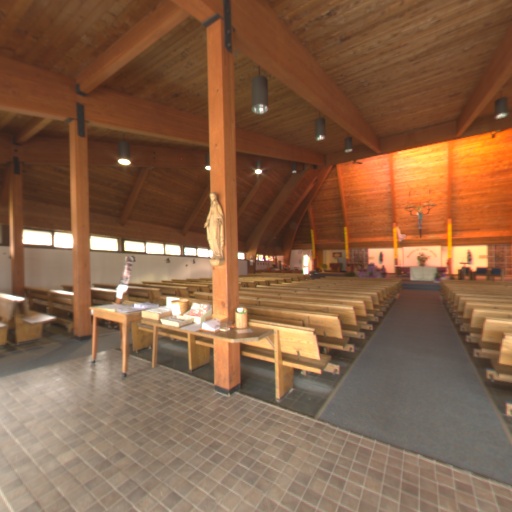}\put(6,10){\color{white}\bfseries\scriptsize -0.522}\end{overpic} \\[2pt]

\vspace{-1.5cm}\rotatebox[origin=c]{90}{\scriptsize{Image $\to$ Text}} & \begin{overpic}[width=\linewidth]{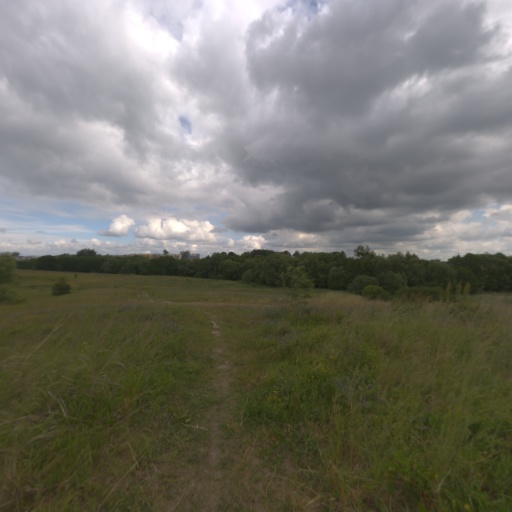}\end{overpic} &
\begin{minipage}[c]{\linewidth}\centering\vspace{-5mm}\scriptsize{\textit{``outdoor, diffused sunlight, overcast sky, soft lighting, muted colors.''}}\\[0.5cm]\textbf{0.998}\end{minipage} &
\begin{minipage}[c]{\linewidth}\centering\vspace{-5mm}\scriptsize{\textit{``outdoor, soft diffused light, cool hue, overcast sky, muted colors, grays and earth tones.''}}\\[0.2cm]\textbf{0.997}\end{minipage} &
\begin{minipage}[c]{\linewidth}\centering\vspace{-5mm}\scriptsize{\textit{``outdoor, soft diffused light, cool muted hue, overcast sky, uniform illumination, subdued grayish tones.''}}\\[0.2cm]\textbf{0.993}\end{minipage} &
\begin{minipage}[c]{\linewidth}\centering\vspace{-5mm}\scriptsize{\textit{``indoor, natural light from tall windows on the right, warm yellow hue, soft illumination, serene atmosphere.''}}\\[0.2cm]\textbf{-0.589}\end{minipage} &
\begin{minipage}[c]{\linewidth}\centering\vspace{-5mm}\scriptsize{\textit{``indoor, natural light from large windows, soft shadows, recessed strip lights, warm inviting tone.''}}\\[0.2cm]\textbf{-0.513}\end{minipage} &
\begin{minipage}[c]{\linewidth}\centering\vspace{-5mm}\scriptsize{\textit{``outdoor, bright sun from the upper right, warm golden hue, di\-rect illumination, natural am\-biance, warm tones dominate.''}}\\[0.02cm]\textbf{-0.512}\end{minipage} \\[2pt]

 & \centering\textbf{Query} & \centering\textbf{Top 1} & \centering\textbf{Top 2} & \centering\textbf{Top 3} & \centering\textbf{Worst 1} & \centering\textbf{Worst 2} & \centering\textbf{Worst 3}
\end{tabular}

    \caption{
        Top-three retrieval comparisons across different modalities in the rows. For example, the first row embeds an image and retrieves environment maps, using the image's \method embedding as a query. Columns show the three best and worst matches. The inset values are the cosine similarities between query and result embeddings.
    }
\label{fig:retrieval_result}
\end{figure*}

Qualitative results of this retrieval task are shown in \cref{fig:retrieval_result}. From the query in the leftmost column, we retrieve the elements from another modality and show the top-3 and bottom-3 matching illumination. We observe that the lighting direction and color are very similar across the top retrieved results, regardless of scene content, highlighting the focus of our representation on lighting. 

We report standard retrieval metrics in \cref{fig:retrieval_stats}, including Recall@K (R@K, whether the correct match appears among the top K results), Mean Reciprocal Rank (MRR), and the median ranks of the ground truth match. Evaluation is performed bidirectionally for each modality pair. The ``Est. Env. map'' modality corresponds to an environment map estimated from the image using DiffusionLight-Turbo \cite{Chinchuthakun2025DiffusionLightTurbo}; as an estimation, it naturally yields lower similarity scores. The recall rate for text is also lower than that of other modalities, likely due to the inherent ambiguity of text descriptions. Despite these differences, similarity scores and ranks across modalities indicate strong cross-modal alignment, effectively translating lighting concepts between environment maps, images, irradiance, and text. As shown in \cref{tab:ablation_retrieval_metrics}, our lighting-specific embedding achieves markedly higher performance than the 2B-parameter embedding version of Qwen3-VL~\cite{qwen3vlembedding} on all retrieval tasks, and CLIP~\cite{radford2021clip} on image$\leftrightarrow$text retrieval across all configurations, demonstrating its effectiveness for lighting-related tasks.

\begin{figure}[bt]
    \centering
    \includegraphics[width=\linewidth]{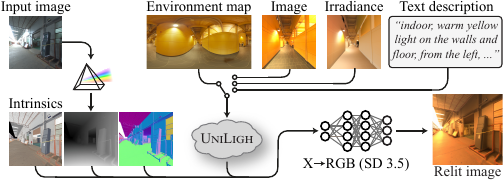}
    \caption{
        Image relighting pipeline using our X$\rightarrow$RGB model. 
    }
    \label{fig:light_control_pipeline}
\end{figure}

\begin{figure}
\resizebox{\columnwidth}{!}{
\footnotesize
\setlength{\tabcolsep}{1pt} 
\renewcommand{\arraystretch}{1} 

\newcommand{\tighttiny}[1]{\begingroup\linespread{0.8}\selectfont\tiny #1\endgroup}

\begin{tabular}{
    m{0.015\textwidth}
    m{0.09\textwidth}
    m{0.09\textwidth}
    m{0.09\textwidth}
    m{0.09\textwidth}
    m{0.09\textwidth}
}
\vspace{-1.5cm}\rotatebox[origin=c]{90}{Image} &
\includegraphics[width=0.09\textwidth]{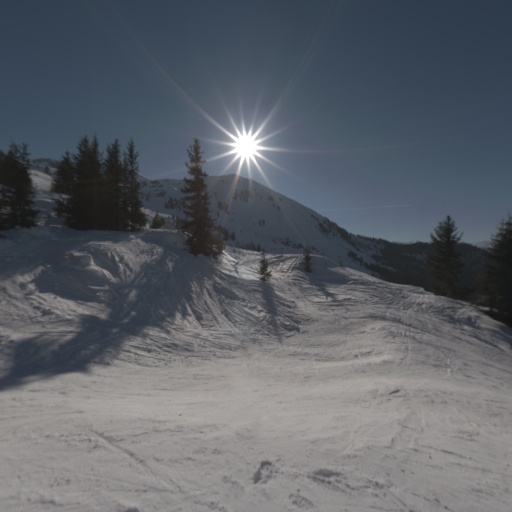} & \includegraphics[width=0.09\textwidth]{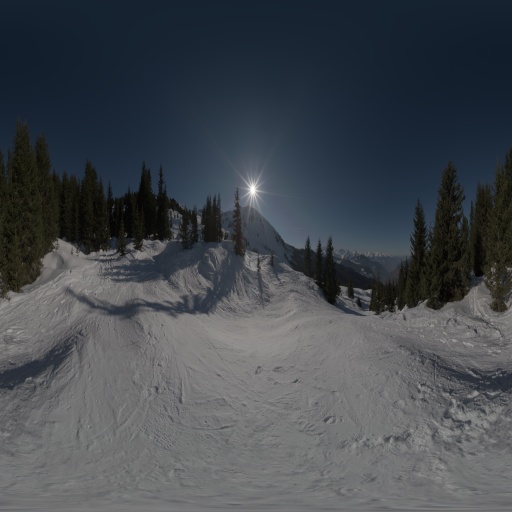} & \includegraphics[width=0.09\textwidth]{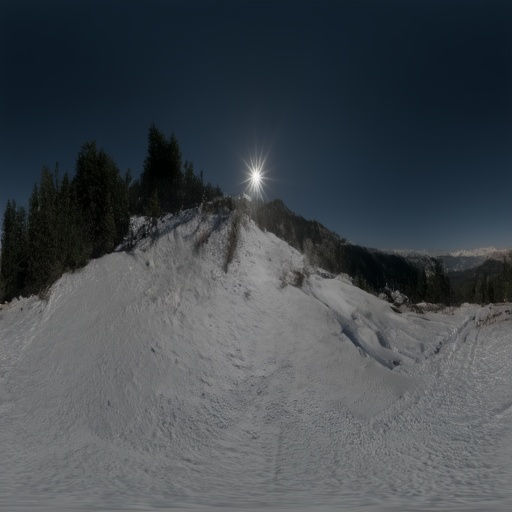} & \includegraphics[width=0.09\textwidth]{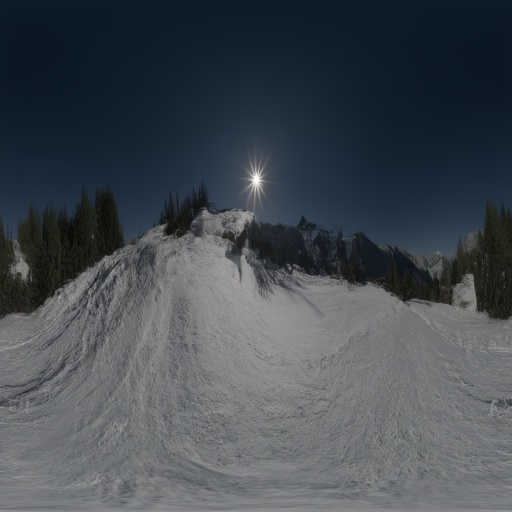} & \includegraphics[width=0.09\textwidth]{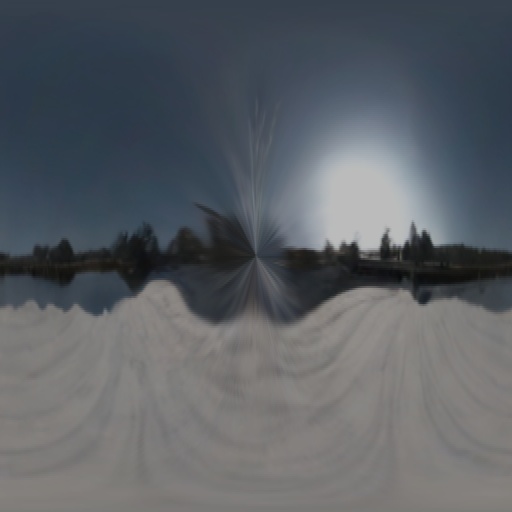} \\[-2pt]
\vspace{-1.5cm}\rotatebox[origin=c]{90}{Env.\,map} &
\includegraphics[width=0.09\textwidth]{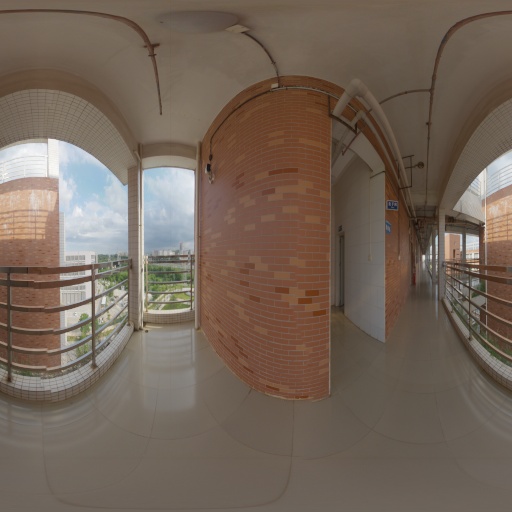} & \includegraphics[width=0.09\textwidth]{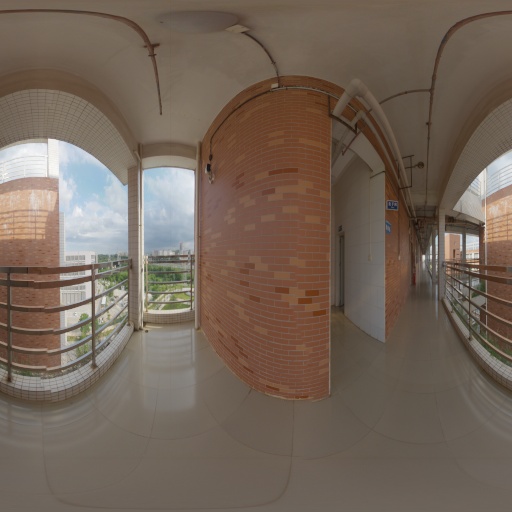} & \includegraphics[width=0.09\textwidth]{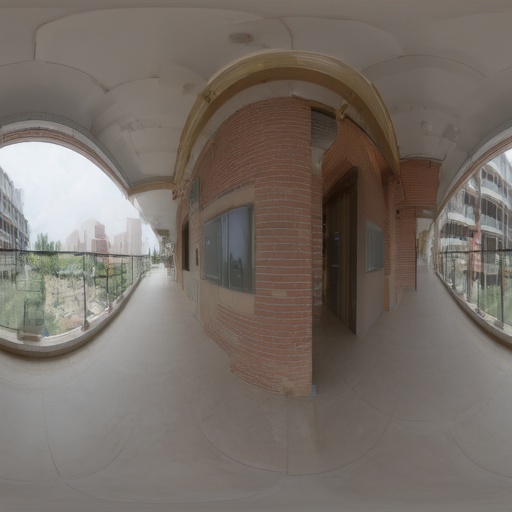} & \includegraphics[width=0.09\textwidth]{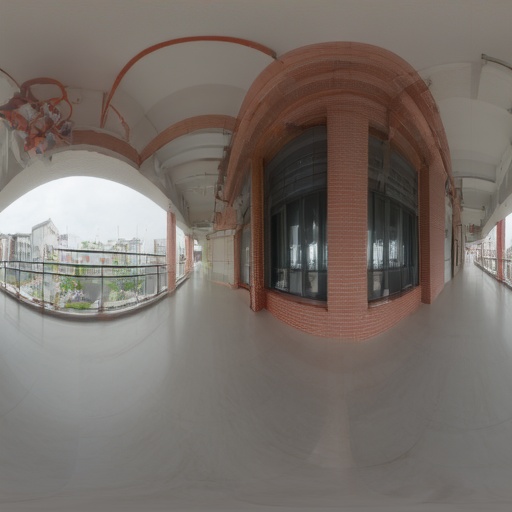} & \multicolumn{1}{c}{N/A} \\[-2pt]
\vspace{-1.5cm}\rotatebox[origin=c]{90}{Text} &
\tiny\textit{``The primary natural light source is the sun, positioned to the upper right, casting a soft and diffused light with a cool blue hue. There are no visible artificial lights.''} & \includegraphics[width=0.09\textwidth]{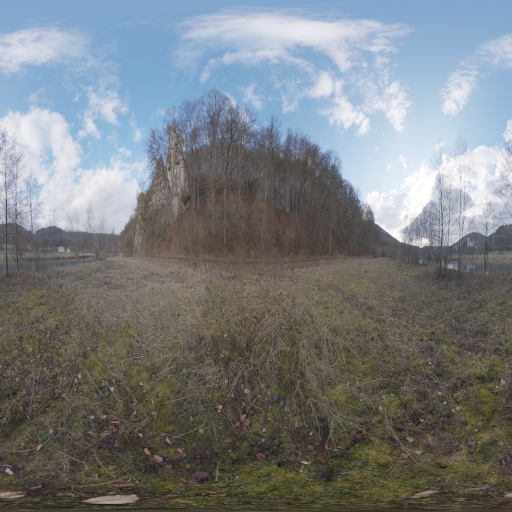} & \includegraphics[width=0.09\textwidth]{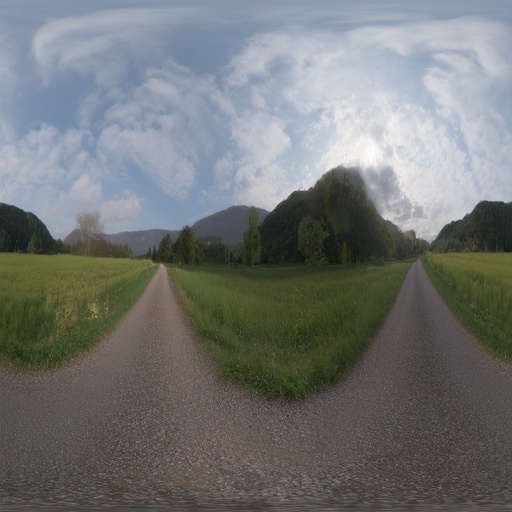} & \includegraphics[width=0.09\textwidth]{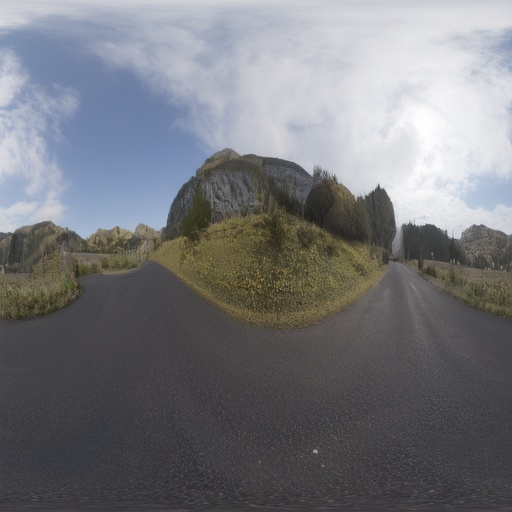} & \multicolumn{1}{c}{N/A} \\[-2pt]
\vspace{-1.5cm}\rotatebox[origin=c]{90}{Irradiance} &
\includegraphics[width=0.09\textwidth]{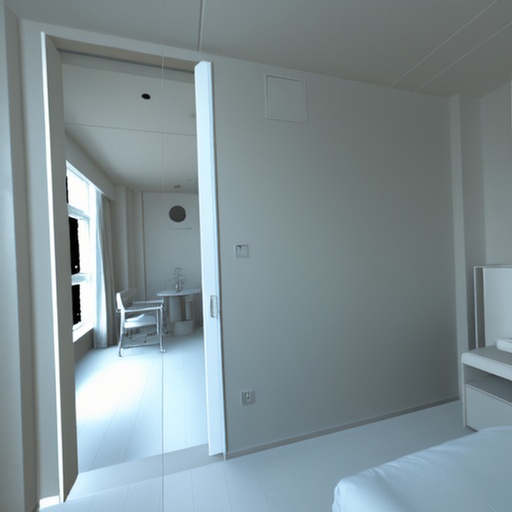} & \includegraphics[width=0.09\textwidth]{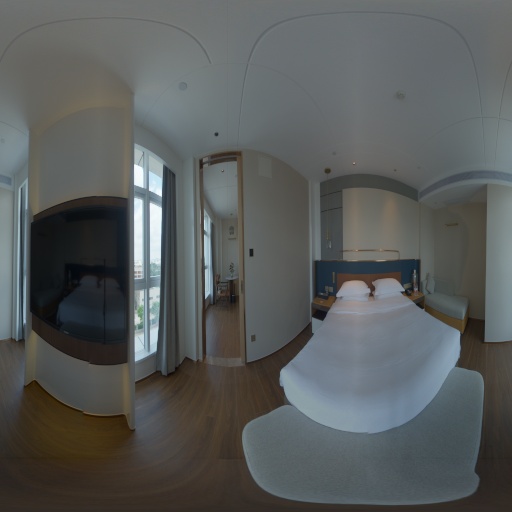} & \includegraphics[width=0.09\textwidth]{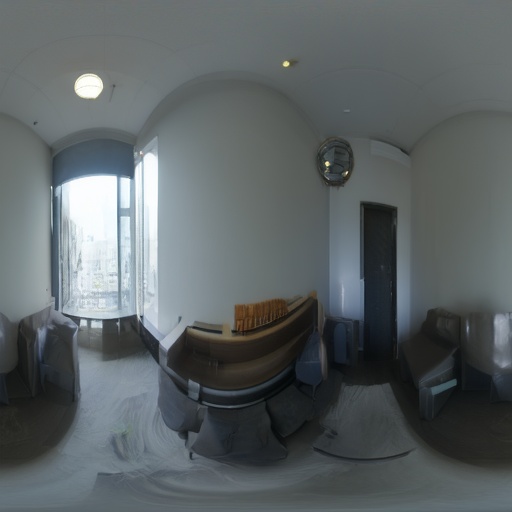} & \includegraphics[width=0.09\textwidth]{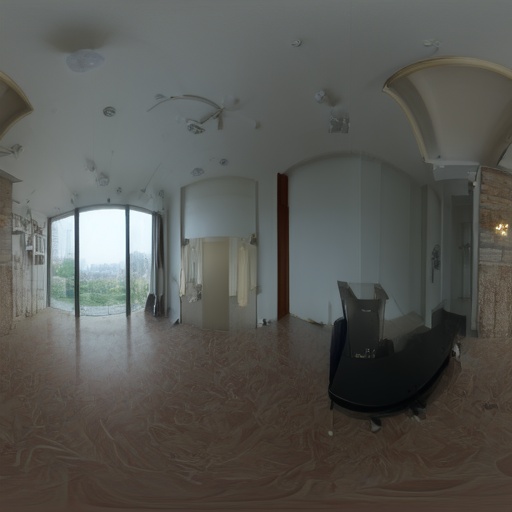} & \multicolumn{1}{c}{N/A} \\[-1pt]

\centering & \centering \!Input modality\! & \centering Reference & \centering\textbf{Ours}\,\scriptsize{(seed 1)} & \centering \textbf{Ours}\,\scriptsize{(seed 2)} & \centering DL-Turbo~\cite{Chinchuthakun2025DiffusionLightTurbo} \\
\end{tabular}
}
    \caption{
        Environment-map generation using our \method-conditioned diffusion model with four control modalities, for two different seeds. We compare to DiffusionLight-Turbo~\cite{Chinchuthakun2025DiffusionLightTurbo} which supports only image conditioning.
    }
    \label{fig:envmap_generation}
\end{figure}

\begin{figure}
\begin{center}
    \resizebox{\columnwidth}{!}{
    \hspace{-4mm}
    \footnotesize
    \setlength{\tabcolsep}{1pt} 
    \renewcommand{\arraystretch}{1} 
    
    \newcommand{\tighttiny}[1]{\begingroup\linespread{0.8}\selectfont\tiny #1\endgroup}
    
    \begin{tabular}{
        m{0.09\textwidth}
        m{0.09\textwidth}
        m{0.09\textwidth}
        m{0.09\textwidth}
        m{0.09\textwidth}
        m{0.09\textwidth}
        m{0.09\textwidth}
    }
    \includegraphics[width=0.09\textwidth]{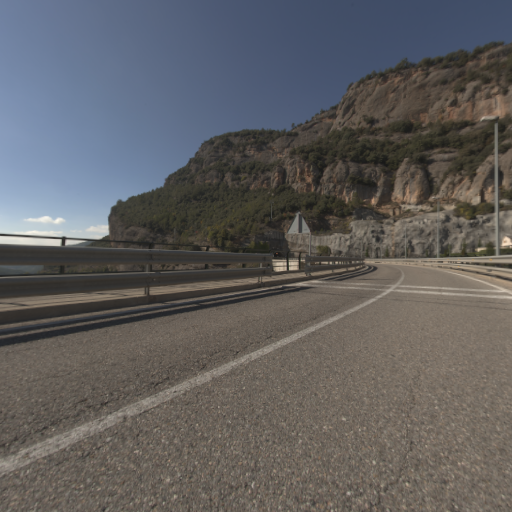} &
    \includegraphics[width=0.09\textwidth]{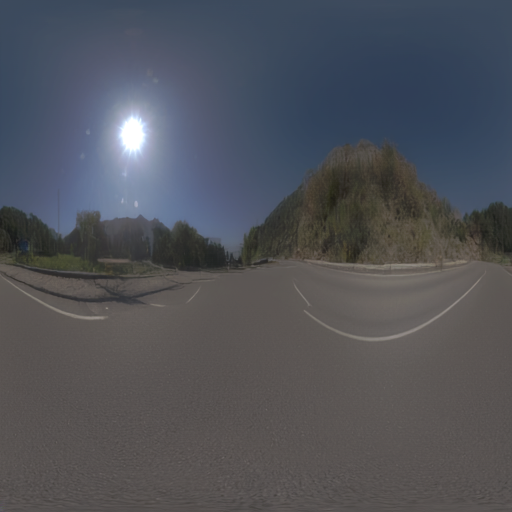} &
    \includegraphics[width=0.09\textwidth]{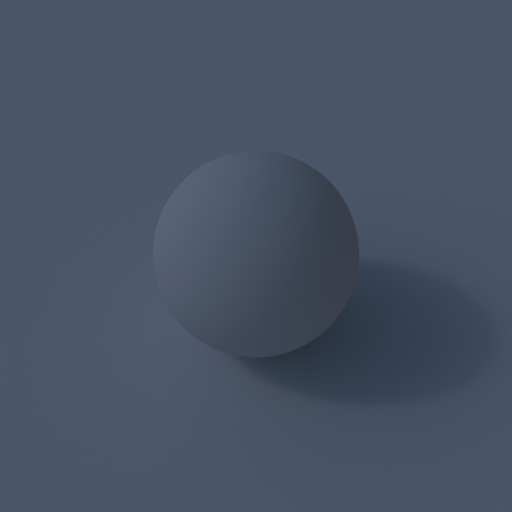} &
    \includegraphics[width=0.09\textwidth]{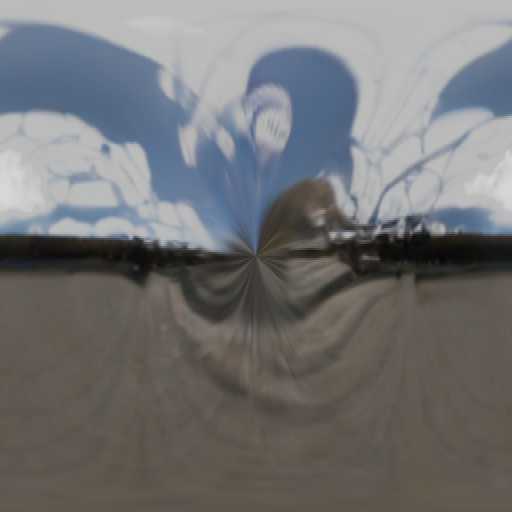} &
    \includegraphics[width=0.09\textwidth]{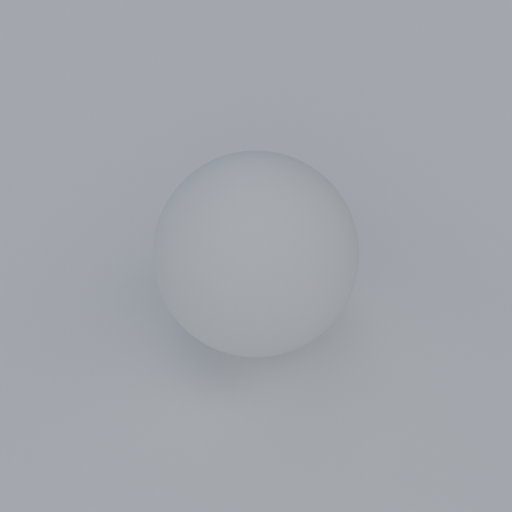} &
    \includegraphics[width=0.09\textwidth]{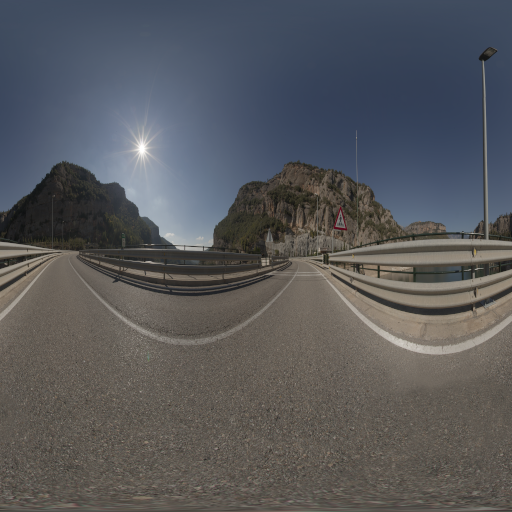} &
    \includegraphics[width=0.09\textwidth]{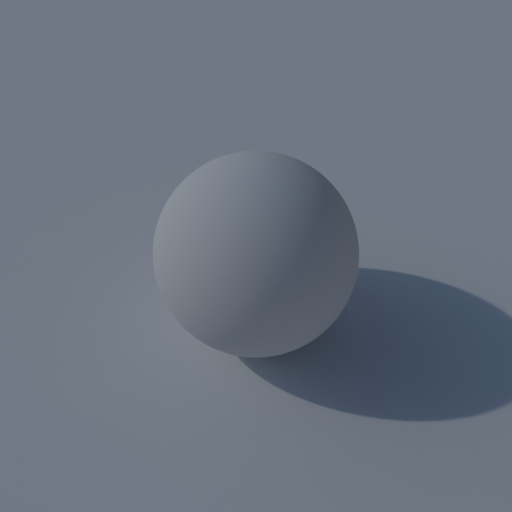} \\[-2pt]
    \includegraphics[width=0.09\textwidth]{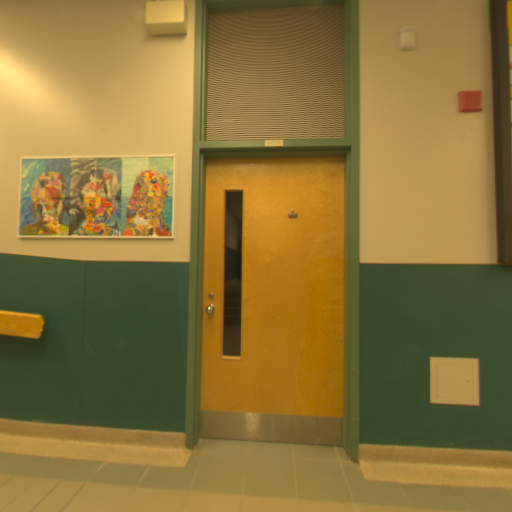} &
    \includegraphics[width=0.09\textwidth]{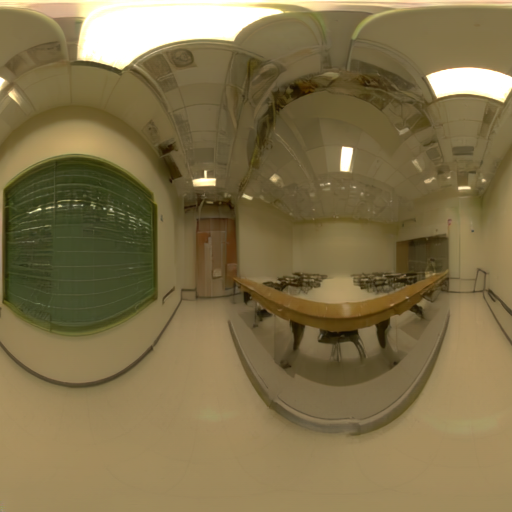} &
    \includegraphics[width=0.09\textwidth]{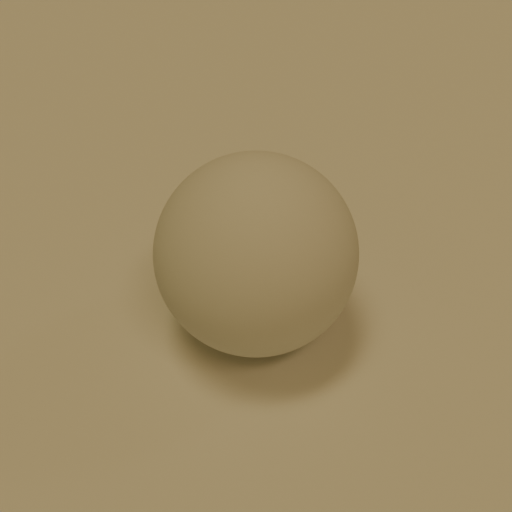} &
    \includegraphics[width=0.09\textwidth]{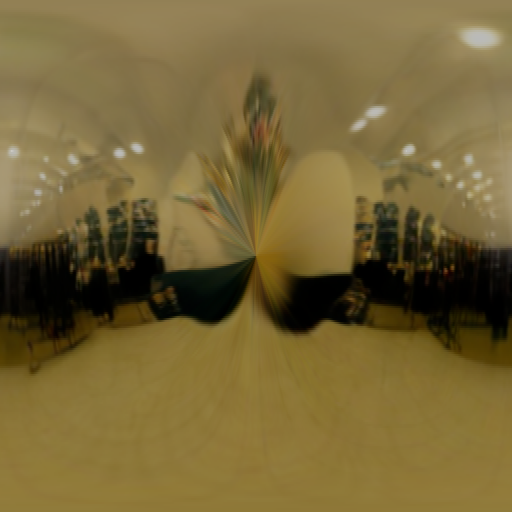} &
    \includegraphics[width=0.09\textwidth]{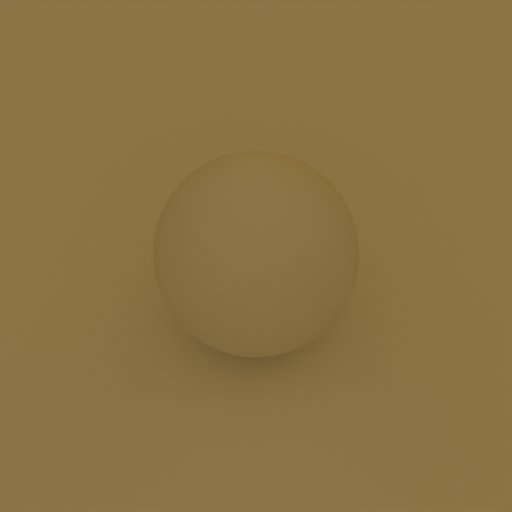} &
    \includegraphics[width=0.09\textwidth]{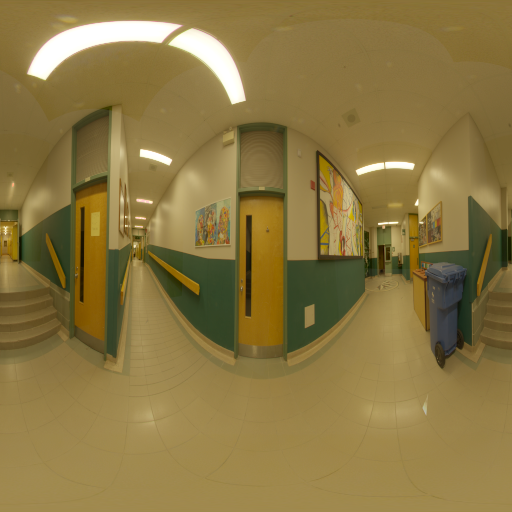} &
    \includegraphics[width=0.09\textwidth]{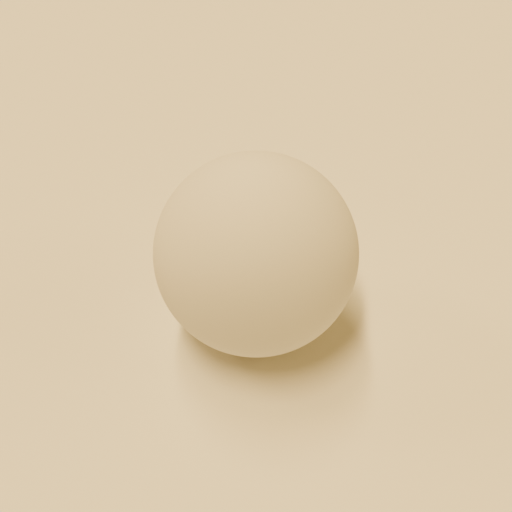}  \\[-1pt]

     \centering Input image & \multicolumn{2}{c}{UniLight} & \multicolumn{2}{c}{DiffusionLight-Turbo} & \multicolumn{2}{c}{Ground truth} \\
    \end{tabular}
    }
\end{center}
    \vspace{-10pt}
    \caption{
        HDR environment-map estimations and their corresponding renders.
    }
    \label{fig:hdr_envmap_render}
    \vspace{-15pt}
\end{figure}

\begin{table}
    \caption{Metrics on the renders of HDR env. map estimations.}
    \label{tab:hdr_envmap_render_metrics}
    \vspace{-15pt} 
    \begin{center}
        \scriptsize
        \setlength{\tabcolsep}{2pt}
        \resizebox{1.0\columnwidth}{!}{
        \begin{tabular}{lcccccc}
            \toprule
            \textbf{Method} & \textbf{PSNR}$_\uparrow$ & \textbf{RMSE}$_\downarrow$ & \textbf{SI-RMSE}$_\downarrow$ & \textbf{SSIM}$_\uparrow$ & \textbf{MAE}$_\downarrow$ & \textbf{LPIPS}$_\downarrow$ \\
            \midrule
            DiffusionLight-Turbo & 27.77 & 0.157 & 0.062 & 0.902 & 0.148 & 0.088 \\
            UniLight & \textbf{28.85} & \textbf{0.133} & \textbf{0.060} & \textbf{0.915} & \textbf{0.124} & \textbf{0.079} \\
            \bottomrule
        \end{tabular}
        }
    \end{center}
    \vspace{-15pt}
\end{table}

\subsection{Environment-map generation}

We fine-tune Stable Diffusion 3.5 Medium~\cite{SD3} to generate $360\degree$ \ac{LDR} environment maps, repurposing its text-conditioning branch to accept our lighting embedding. This approach produces structurally coherent maps consistent with diverse input modalities (\cref{fig:envmap_generation}). To achieve \ac{HDR}, we employ a separate multi-exposure reconstruction model. As shown in \cref{fig:hdr_envmap_render} and \cref{tab:hdr_envmap_render_metrics}, our renders significantly outperform DiffusionLight-Turbo~\cite{Phongthawee2024DiffusionLight, Chinchuthakun2025DiffusionLightTurbo} in both visual quality and quantitative metrics. A detailed pipeline diagram is included in the supplementary.


\begin{figure*}[t]
    \centering
    \includegraphics{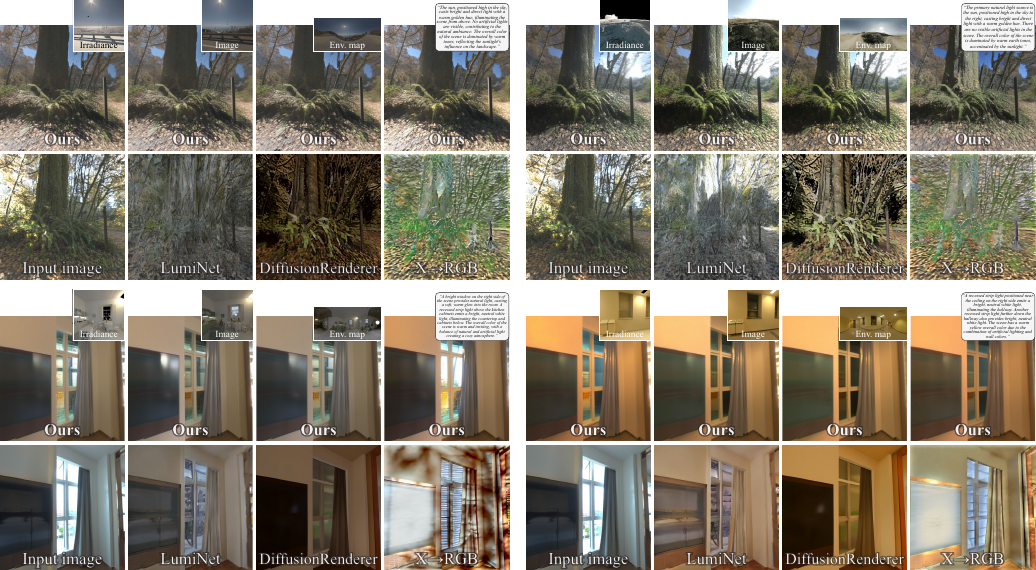}%
    \vspace{-1.5mm}%
    \caption{
        Image relighting using our \method-conditioned X$\rightarrow$RGB model (see \cref{fig:light_control_pipeline}) with different lighting modalities. We apply four lighting conditions on two images (rows) and compare against three baselines (columns): LumiNet~\cite{xing2025luminet} (image modality), DiffusionRenderer~\cite{DiffusionRenderer} (environment map), and original X$\rightarrow$RGB~\cite{zeng2024rgbx} (text). We also show irradiance-conditioned generation with our model; comparing to original X$\rightarrow$RGB~\cite{zeng2024rgbx} here is moot as that model requires the irradiance to be aligned with (the intrinsics of) the input image.
    }
    \vspace{-2.5mm}
    \label{fig:light_control_combined2}
\end{figure*}

\begin{figure}

\begin{center}
    \centering
    \resizebox{1.0\columnwidth}{!}{
        \footnotesize
        \setlength{\tabcolsep}{1pt}
        \renewcommand{\arraystretch}{1}
        \hspace{-3mm}
        \begin{tabular}{
            m{0.01\textwidth}
            m{0.09\textwidth}
            m{0.09\textwidth}
            m{0.09\textwidth}
            m{0.09\textwidth}
            m{0.09\textwidth}
            m{0.09\textwidth}
        }
            & \scriptsize \centering Env.\,map
            & \scriptsize \centering Qwen3-VL
            & \scriptsize \centering \textbf{Ours}
            & \scriptsize \centering Env.\,map
            & \scriptsize \centering  Qwen3-VL
            & \scriptsize ~~~~~~~~\textbf{Ours} \\
            %
            \vspace{-1.5cm}\rotatebox[origin=c]{90}{\scriptsize{-120\degree}}

            & \includegraphics[width=0.09\textwidth]{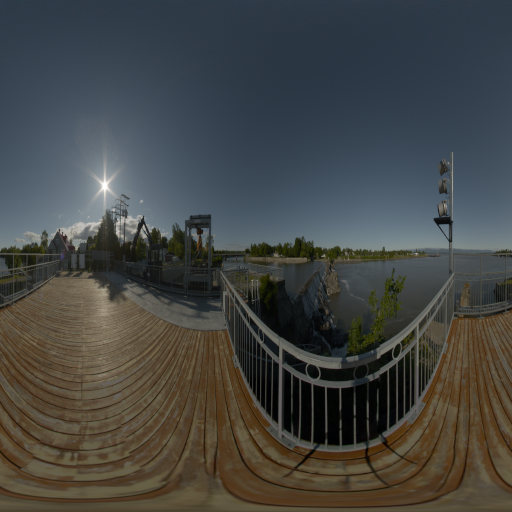}
            & \includegraphics[width=0.09\textwidth]{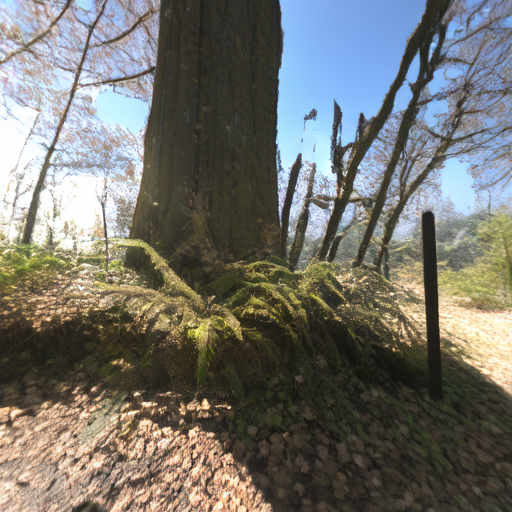}
            & \includegraphics[width=0.09\textwidth]{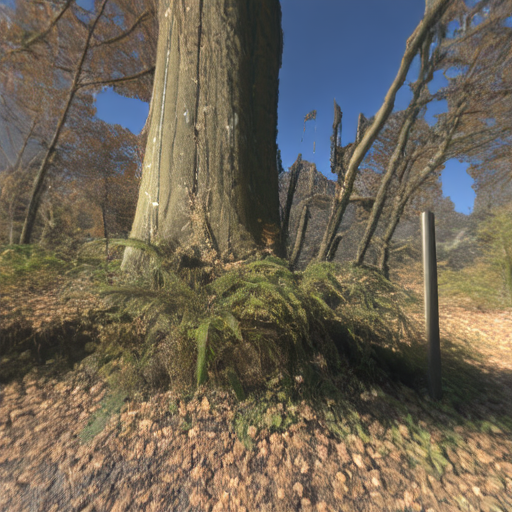}

            & \includegraphics[width=0.09\textwidth]{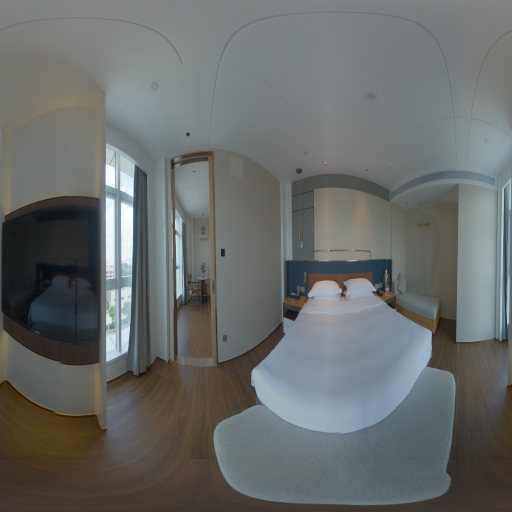}
            & \includegraphics[width=0.09\textwidth]{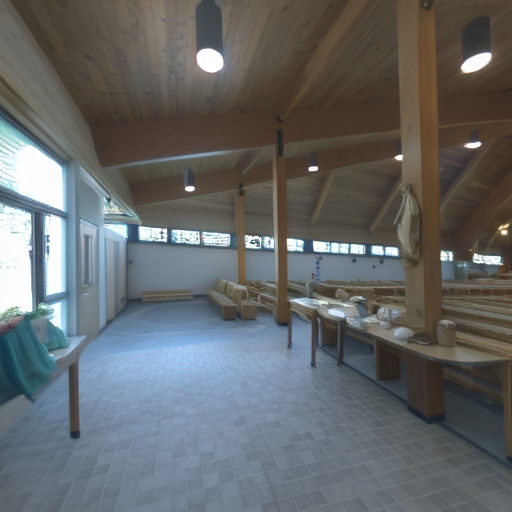}
            & \includegraphics[width=0.09\textwidth]{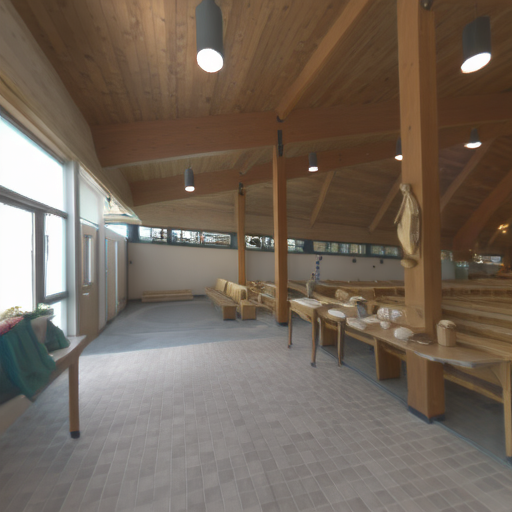}
            \\[4pt]

            \vspace{-1.5cm}\rotatebox[origin=c]{90}{\scriptsize{-60\degree}}

            & \includegraphics[width=0.09\textwidth]{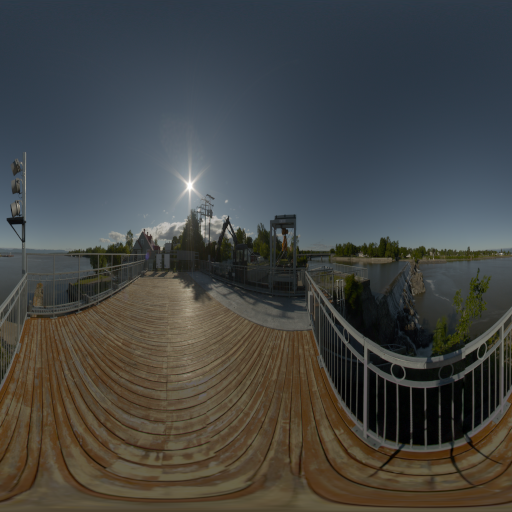}
            & \includegraphics[width=0.09\textwidth]{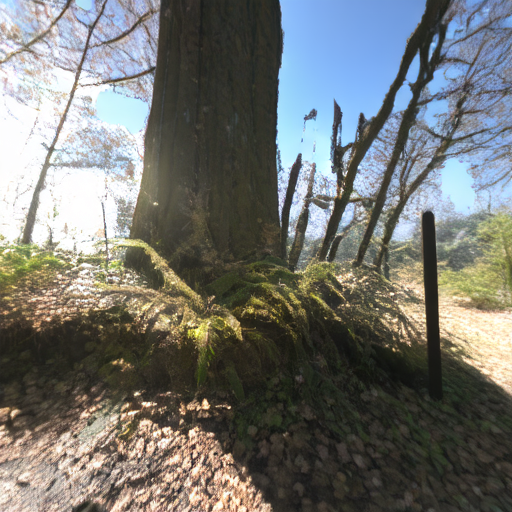}
            & \includegraphics[width=0.09\textwidth]{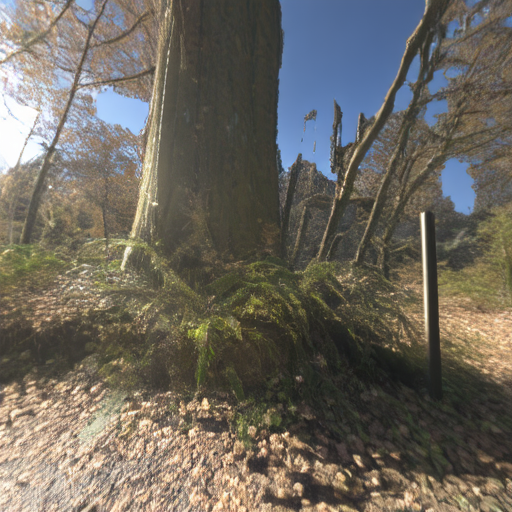}

            & \includegraphics[width=0.09\textwidth]{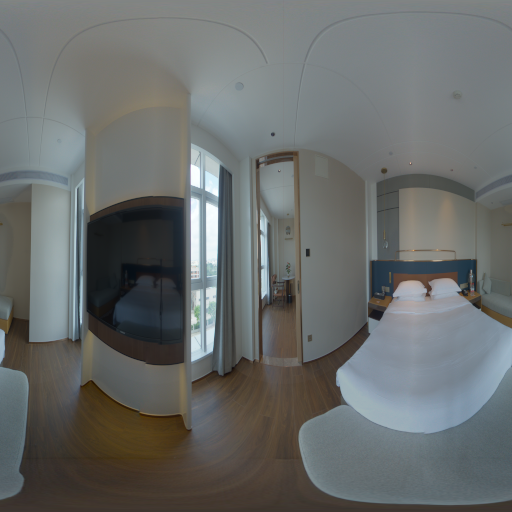}
            & \includegraphics[width=0.09\textwidth]{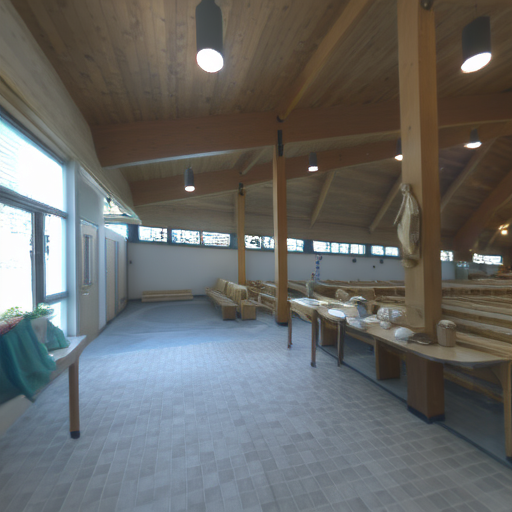}
            & \includegraphics[width=0.09\textwidth]{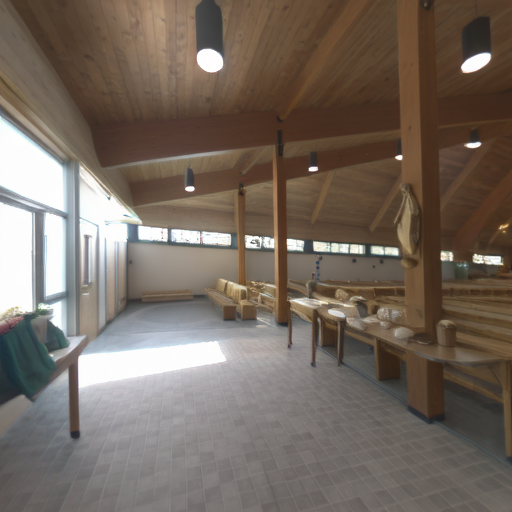}
            \\[4pt]

            \vspace{-1.5cm}\rotatebox[origin=c]{90}{\scriptsize{0\degree}}

            & \includegraphics[width=0.09\textwidth]{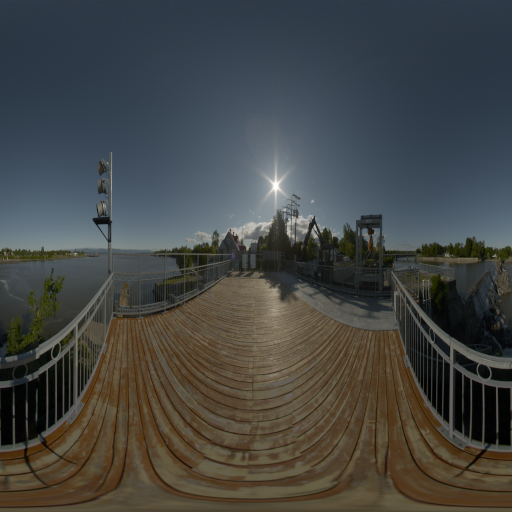}
            & \includegraphics[width=0.09\textwidth]{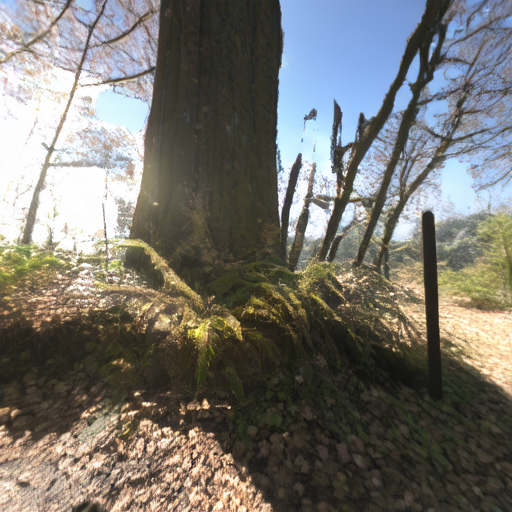}
            & \includegraphics[width=0.09\textwidth]{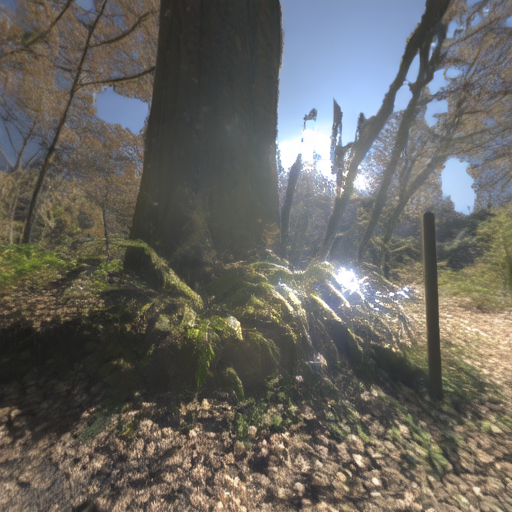}

            & \includegraphics[width=0.09\textwidth]{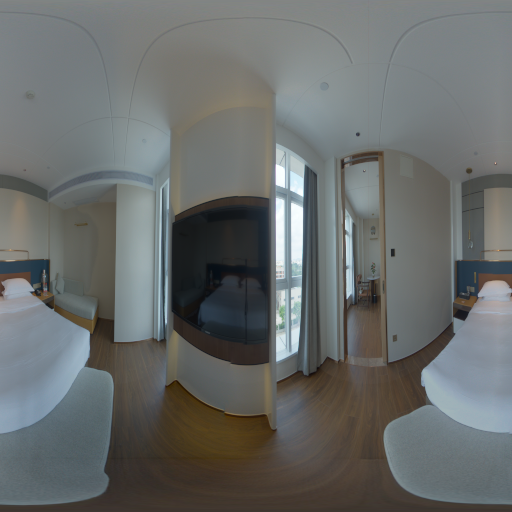}
            & \includegraphics[width=0.09\textwidth]{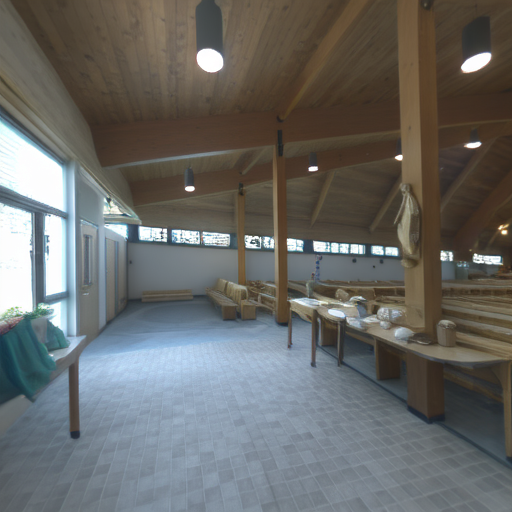}
            & \includegraphics[width=0.09\textwidth]{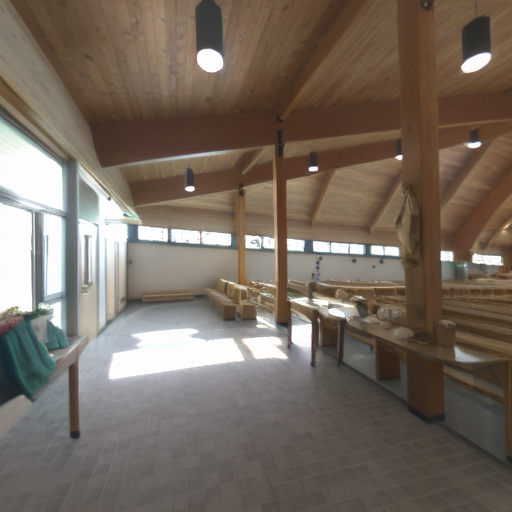}
            \\[4pt]

            \vspace{-1.5cm}\rotatebox[origin=c]{90}{\scriptsize{+60\degree}}

            & \includegraphics[width=0.09\textwidth]{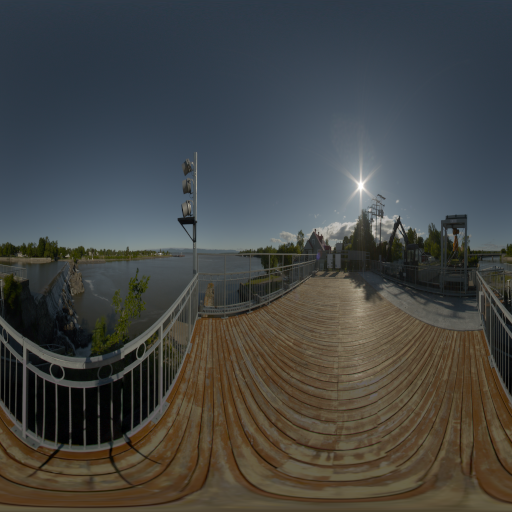}
            & \includegraphics[width=0.09\textwidth]{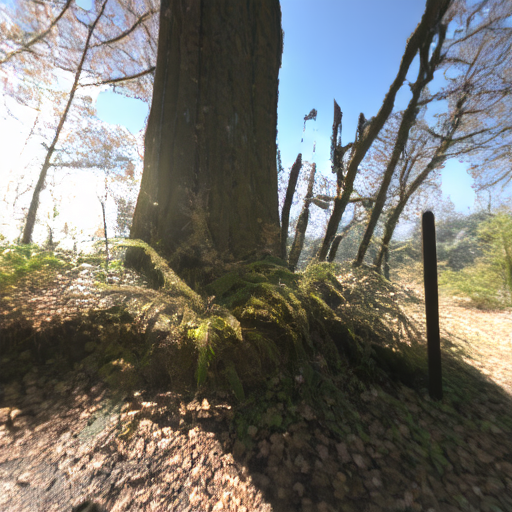}
            & \includegraphics[width=0.09\textwidth]{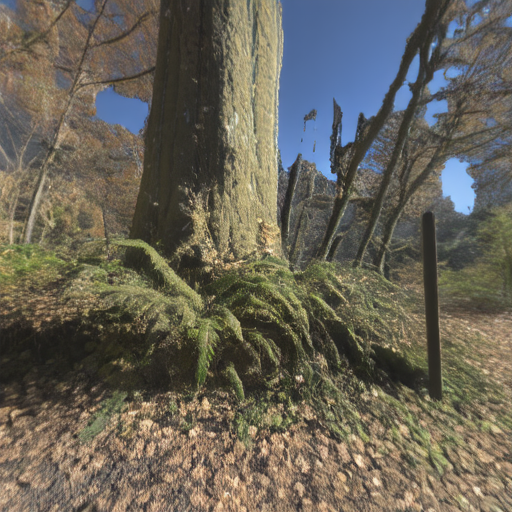}

            & \includegraphics[width=0.09\textwidth]{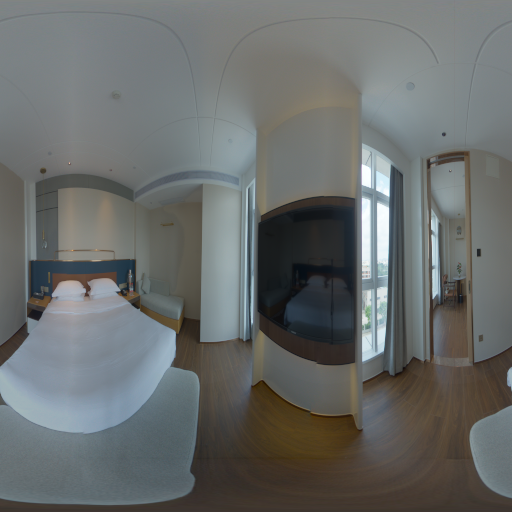}
            & \includegraphics[width=0.09\textwidth]{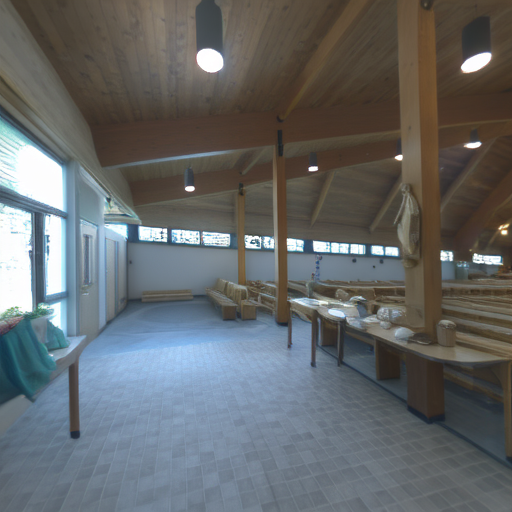}
            & \includegraphics[width=0.09\textwidth]{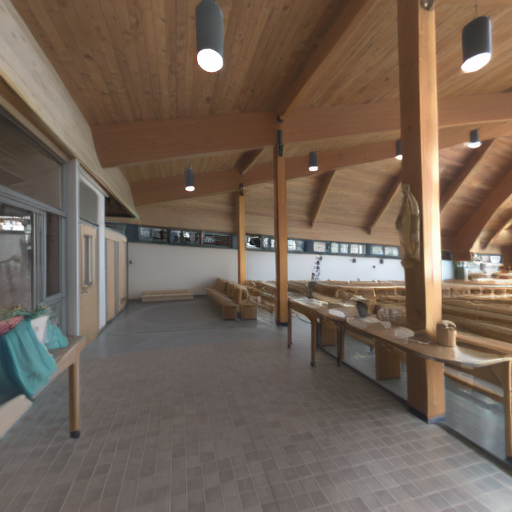}
            \\[4pt]

            \vspace{-1.5cm}\rotatebox[origin=c]{90}{\scriptsize{+120\degree}}

            & \includegraphics[width=0.09\textwidth]{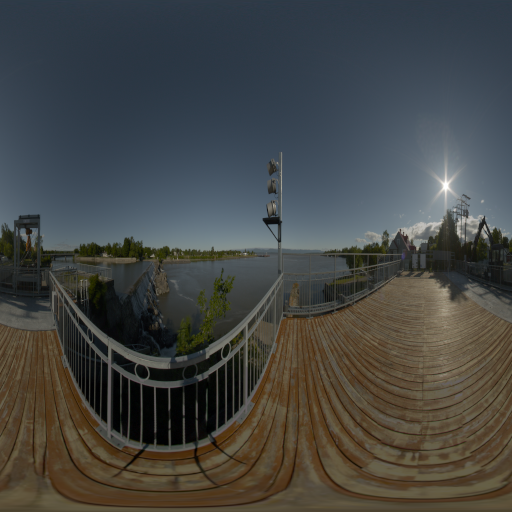}
            & \includegraphics[width=0.09\textwidth]{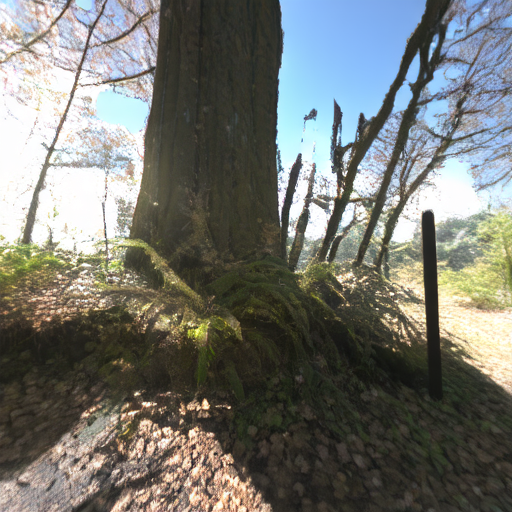}
            & \includegraphics[width=0.09\textwidth]{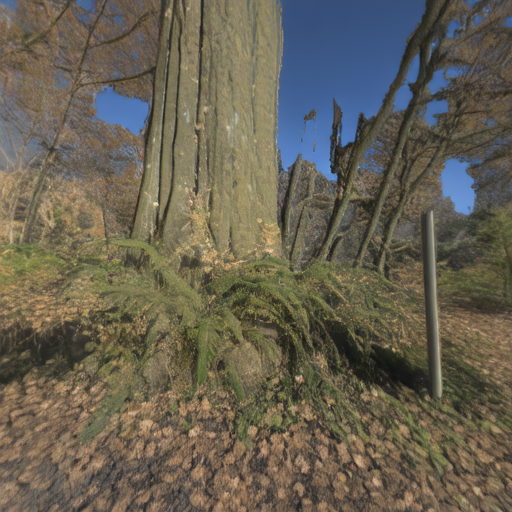}

            & \includegraphics[width=0.09\textwidth]{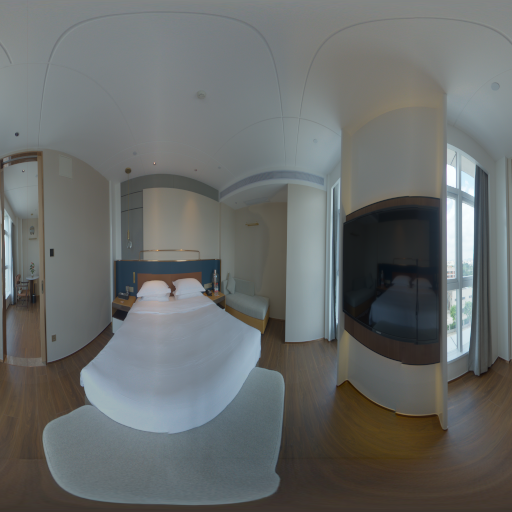}
            & \includegraphics[width=0.09\textwidth]{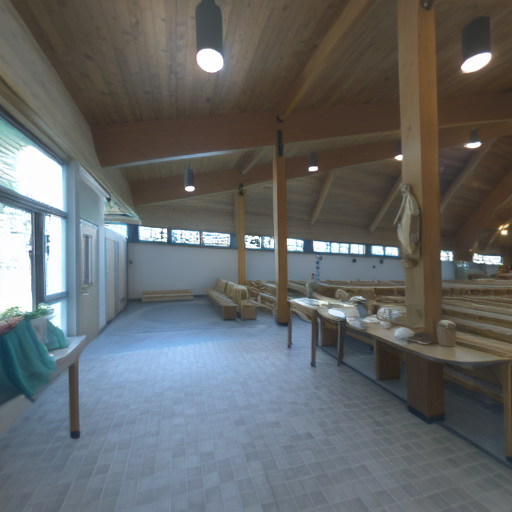}
            & \includegraphics[width=0.09\textwidth]{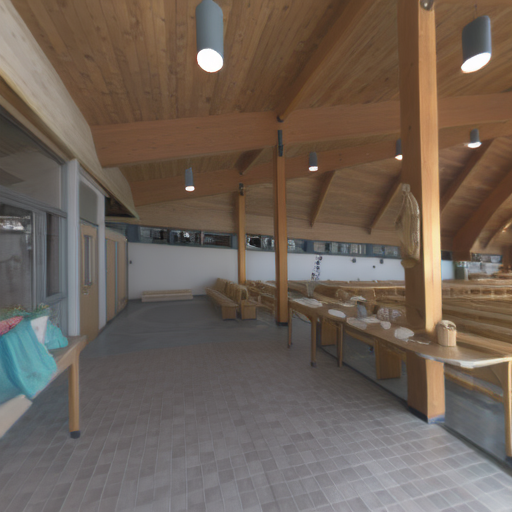}
        \end{tabular}
    }
\end{center}

    \vspace{-6mm}
    \caption{
        Comparison on image relighting using Qwen3-VL and our embeddings, with env.\ map rotated in $60\degree$ intervals.
    }
    \label{fig:light_control_envmap_rotation}
\end{figure}

\subsection{Light control for image generation}

To assess the utility of our representation for relighting, we integrate it into the X$\rightarrow$RGB framework~\cite{zeng2024rgbx} based on Stable Diffusion 3.5 Medium~\cite{SD3}, fine-tuned in two stages. First, a base X$\rightarrow$RGB model is trained to generate renders from intrinsics (depth, normal, albedo, and irradiance), using InteriorVerse~\cite{zhu2022learning}, Hypersim~\cite{roberts2021hypersim}, our own synthetic data, and real images with estimated intrinsics. Second, we repurpose the text-conditioning branch to accept our \method embeddings instead of prompts, fine-tuning on our multi-modal dataset to reconstruct images from intrinsics with zeroed irradiance (\cref{fig:light_control_pipeline}). While alternatives like Qwen3-VL~\cite{qwen3vlembedding} capture general semantics, they struggle with precise lighting control, often leaving shadows and highlights static during environment-map rotation (\cref{fig:light_control_envmap_rotation}). For outdoor scenes, we apply a sky masking model~\cite{xie2021segformer} to ensure lighting coherence. Our approach significantly outperforms baselines qualitatively (\cref{fig:light_control_combined2}).

\section{Discussion}

Our work represents a step toward a unified representation of lighting, demonstrating a shared latent space that bridges different modalities. Despite this advance, some limitations remain.
First, although our current representation encodes light direction effectively, it does not represent the light spatial variation, which can be important for complex indoor scenes. This could be improved by attempting to merge more spatial modalities \cite{magar2025lightlab} for local control and our supported high-level representations (e.g., environment map, text) for global control. More generally, including additional representations (e.g., shading-based~\cite{fortier2026spotlight}) to further facilitate user interaction is an interesting avenue for future work.
Finally, aligning our embeddings to existing popular encoders~\cite{yu2024representation} may simplify the diffusion model fine-tuning using our lighting embeddings or enable zero-shot use.


\section{Conclusion}

We presented \method, a unified latent lighting representation that operates across multiple modalities, including text, images, environment maps, and irradiance. Our latent space is learned through a contrastive framework grounded with a spherical harmonics loss. We demonstrated the accuracy of our representation through multi-modal light retrieval and showcased its effectiveness in downstream tasks such as light generation and relighting. We believe our contributions lay the groundwork for flexible physically grounded control of generative models, opening new possibilities for lighting-aware synthesis and editing. 

\myparagraph{Acknowledgments.}
This work was partially supported by NSERC grant RGPIN-2020-04799 and partially done during Zhang’s internship at Adobe Research London. Computing resources were provided by Adobe and the Digital Research Alliance of Canada.





{
    \small
    \bibliographystyle{ieeenat_fullname}
    \bibliography{references}
}



\end{document}